\documentclass{article}
\usepackage[utf8]{inputenc}

\usepackage[preprint]{neurips_2026}

\workshoptitle{The 6th Workshop on Mathematical Reasoning and AI}

\usepackage{amssymb}

\usepackage{amsmath}
\usepackage{enumitem}
\usepackage[utf8]{inputenc} 
\usepackage{amssymb} 
\usepackage[T1]{fontenc}    
\usepackage{hyperref}       
\usepackage{xurl}           
\usepackage{booktabs}       
\usepackage{graphicx}       
\usepackage{amsfonts}       
\usepackage{nicefrac}       
\usepackage{microtype}      
\usepackage{xcolor}         
\newcommand{\stochbench}{\textsc{StochBench}}
\usepackage{booktabs}
\usepackage{siunitx}
\usepackage{subcaption}
\usepackage{listings}
\usepackage{amssymb}
\usepackage{booktabs,subcaption}
\usepackage{color}
\definecolor{keywordcolor}{rgb}{0.7, 0.1, 0.1}   
\definecolor{tacticcolor}{rgb}{0.0, 0.1, 0.6}    
\definecolor{commentcolor}{rgb}{0.4, 0.4, 0.4}   
\definecolor{symbolcolor}{rgb}{0.0, 0.1, 0.6}    
\definecolor{sortcolor}{rgb}{0.1, 0.5, 0.1}      
\definecolor{attributecolor}{rgb}{0.7, 0.1, 0.1} 
\usepackage{newunicodechar}
\newunicodechar{ℱ}{\ensuremath{\mathcal{F}}}

\newunicodechar{⨆}{\ensuremath{\bigsqcup}}

\newunicodechar{ᵐ}{\ensuremath{^m}}
\usepackage{soul}

\definecolor{darkblue}{rgb}{0, 0, 0.5}
\hypersetup{colorlinks=true, citecolor=darkblue, linkcolor=darkblue, urlcolor=darkblue}
\lstdefinelanguage{lean}{
  sensitive=true,
  mathescape=false,
  texcl=false,
  morekeywords=[1]{
    import,prelude,protected,private,noncomputable,
    definition,meta,renaming,hiding,parameter,parameters,
    begin,constant,constants,lemma,variable,variables,
    print,theorem,example,open,export,axiom,axioms,
    inductive,with,structure,record,universe,universes,
    match,infix,infixl,infixr,notation,postfix,prefix,
    instance,end,namespace,section,attribute,local,
    set_option,extends,include,omit,class,calc,have,
    show,suffices,by,in,at,let,forall,fun,exists,
    if,then,else,obtain,from,mutual,do,def,partial,
    mut,where,macro,syntax,deriving,return,try,catch,
    for,macro_rules,declare_syntax_cat,abbrev
  },
  morekeywords=[2]{Sort,Type,Prop},
  morekeywords=[3]{
    assumption,apply,intro,intros,generalize,clear,
    revert,done,exact,refine,repeat,cases,rewrite,rw,
    simp,simp_all,contradiction,constructor,injection,
    induction,rcases,rintro,obtain,simpa,norm_num,
    positivity,linarith,nlinarith,ring,ring_nf,
    aesop,ext,funext,congr,convert,filter_upwards
  },
  morecomment=[l]{--},
  morecomment=[n]{/-}{-/},
  morestring=[b]",
  inputencoding=utf8,
  extendedchars=true,
  basicstyle=\ttfamily\footnotesize,
  keywordstyle=[1]{\color{keywordcolor}},
  keywordstyle=[2]{\color{sortcolor}},
  keywordstyle=[3]{\color{tacticcolor}},
  identifierstyle=\color{black},
  commentstyle=\color{commentcolor},
  stringstyle=\ttfamily,
  columns=fullflexible,
  keepspaces=true,
  showstringspaces=false,
  tabsize=3,
  breaklines=true,
  breakatwhitespace=false,
  captionpos=b,
  literate=
    {·}{{\ensuremath{\cdot}}}1
    {×}{{\ensuremath{\times}}}1
    {à}{{\`{a}}}1
    {Ω}{{\ensuremath{\Omega}}}1
    {θ}{{\ensuremath{\theta}}}1
    {μ}{{\ensuremath{\mu}}}1
    {π}{{\ensuremath{\pi}}}1
    {σ}{{\ensuremath{\sigma}}}1
    {φ}{{\ensuremath{\varphi}}}1
    {ω}{{\ensuremath{\omega}}}1
    {ᵐ}{{\ensuremath{{}^{\mathrm{m}}}}}1
    {‖}{{\ensuremath{\Vert}}}1
    {•}{{\ensuremath{\bullet}}}1
    {ℕ}{{\ensuremath{\mathbb{N}}}}1
    {ℝ}{{\ensuremath{\mathbb{R}}}}1
    {ℱ}{{\ensuremath{\mathcal{F}}}}1
    {←}{{\ensuremath{\leftarrow}}}1
    {→}{{\ensuremath{\rightarrow}}}1
    {∀}{{\ensuremath{\forall}}}1
    {∂}{{\ensuremath{\partial}}}1
    {∃}{{\ensuremath{\exists}}}1
    {∈}{{\ensuremath{\in}}}1
    {∑}{{\ensuremath{\sum}}}1
    {∧}{{\ensuremath{\wedge}}}1
    {∫}{{\ensuremath{\int}}}1
    {≠}{{\ensuremath{\ne}}}1
    {≤}{{\ensuremath{\le}}}1
    {≥}{{\ensuremath{\ge}}}1
    {⊢}{{\ensuremath{\vdash}}}1
    {▸}{{\ensuremath{\blacktriangleright}}}1
    {⟨}{{\ensuremath{\langle}}}1
    {⟩}{{\ensuremath{\rangle}}}1
    {⨆}{{\ensuremath{\bigsqcup}}}1
}

\title{\stochbench: A Domain-Specific Benchmark \\ for Stochastic Processes in Lean}

\author{%
  \textbf{Idan Davidovich} \quad
  \textbf{Debargha Ganguly} \quad
  \textbf{Vikash Singh} \quad
  \textbf{Vipin Chaudhary} \\
  Case Western Reserve University \\
  \texttt{\{idan, debargha, vikash, vipin\}@case.edu}\\
   \url{https://huggingface.co/datasets/IdanDavidovich/StochBench}\\
}

\begin{document}

\raggedbottom

\maketitle

\begin{abstract}

Leading benchmarks for formal theorem proving with large language models are small collections drawn from competition math, such as the IMO and Putnam, that poorly represent field-specific applications. We introduce \stochbench{}, a Lean~4 benchmark of 450 graduate stochastic-processes problems at varying abstraction levels, each paired with its natural-language source. Addressing a field underrepresented in Mathlib, it covers finite and countable Markov chains, renewal processes, random walks, martingales, stopping times, queues, Brownian motion, stochastic calculus, weak convergence, and Poisson and continuous-time Markov processes. Our Opus~4.8-based agent achieves a 34.9\% proof rate (157/450) under a 15-minute per-problem limit. \stochbench{} better represents domain-specific applied mathematics while remaining challenging for advanced provers.
\end{abstract}


\section{Introduction}
\label{sec:introduction}

Lean enables machine-checkable mathematics, with substantial formalizations including sphere packing in dimension eight, Brownian motion, and Fermat's Last Theorem for regular primes \citep{hariharan2026milestoneformalizationspherepacking,degenne2025formalizationbrownianmotionlean,best2025completeformalizationfermatstheorem}.
Extending this progress to everyday mathematical assistance requires evaluating how consistently automated provers handle a discipline's recurring arguments.
Competition benchmarks and broad textbook collections offer valuable tests, but aggregate scores can obscure domain-specific strengths and failures \citep{zheng2022minif2f,azerbayev2023proofnet,tsoukalas2024putnambench}.

We introduce \stochbench{}, a Lean~4 benchmark for graduate stochastic processes, a field central to statistics and machine learning.
Concentrating on related problems in Markov chains, martingales, and continuous-time processes, we prioritize within-domain depth over cross-domain breadth.
Some problems, however, require infrastructure unavailable in the Mathlib environment \citep{undergrad_todo}.
\emph{Direct} targets use Mathlib or shared definitions, while \emph{abstracted} targets take the required properties as hypotheses. Lean verifies that each proved conclusion follows from its stated hypotheses. We have taken utmost care to make sure the (all human written) definitions and hypotheses faithfully represent the source problem, but can benefit from further peer review.

Our contributions are:

\begin{enumerate}[nosep]
    \item \textbf{A domain-focused benchmark.}
    We release 450 Lean~4 theorem targets paired with informal statements, alongside shared definitions and baseline proof attempts.

    \item \textbf{A scope-aware baseline evaluation.}
    We annotate formalization scope and evaluate a compiler-guided proof agent under a 15-minute per-problem cap, reporting results by topic and representation.
\end{enumerate}


\begin{figure*}[htbp]
    \centering
    \includegraphics[width=\linewidth]{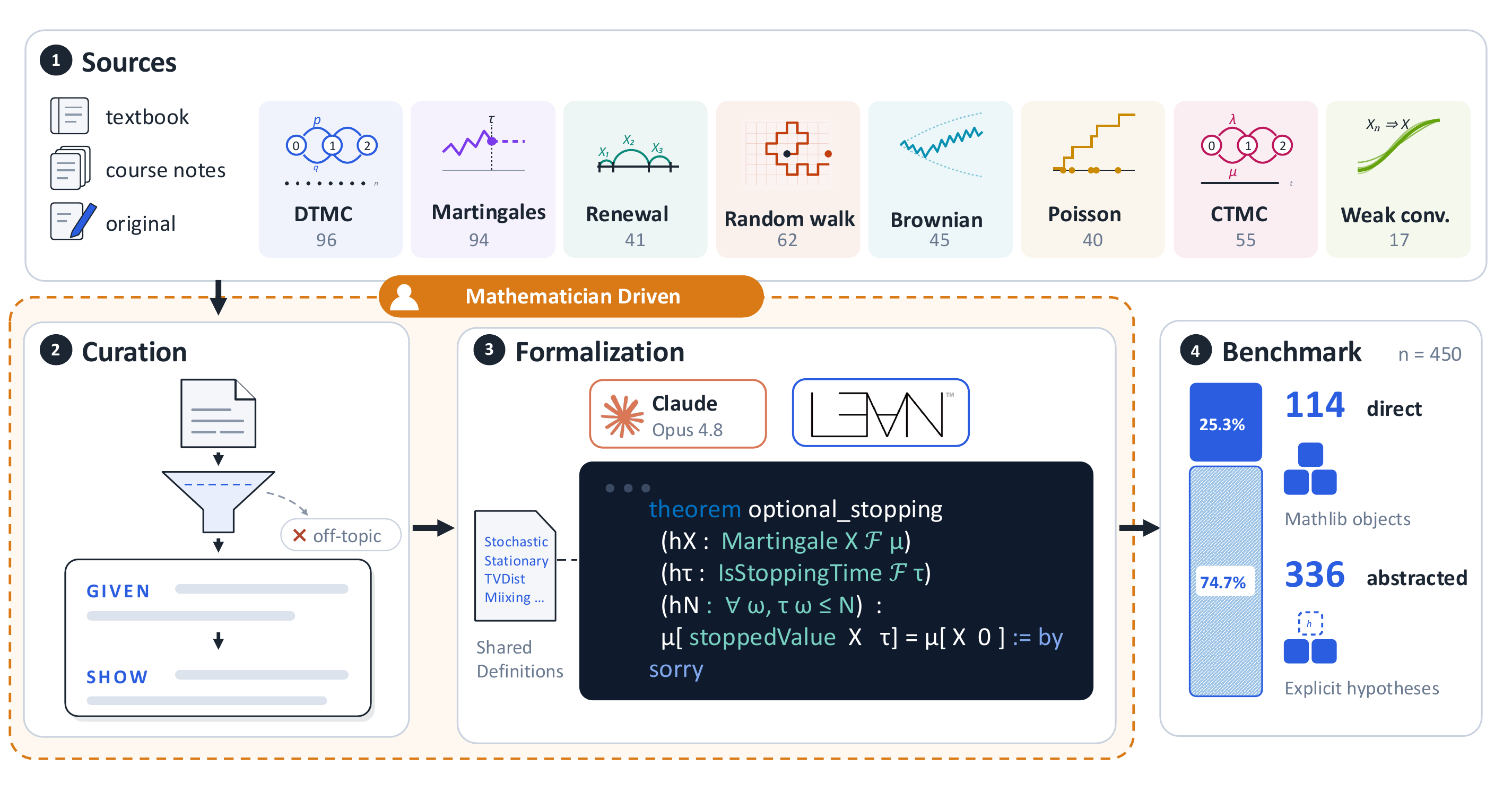}
    \caption{Construction of \stochbench{}: mathematician-led curation and LLM-assisted formalization with shared definitions produce 450 Lean~4 targets across eight topics, comprising 114 direct and 336 abstracted statements.}
    \label{fig:placeholder}
\end{figure*}

\section{Related Work}
\label{sec:related-work}

\textbf{Benchmarks for formal mathematical reasoning.}
Lean-based evaluation has developed along complementary axes of competition difficulty, curricular coverage, and research context.
\textsc{miniF2F} \citep{zheng2022minif2f} established a benchmark centered on Olympiad-style mathematics, \textsc{ProofNet} \citep{azerbayev2023proofnet} paired informal statements and proofs with formal undergraduate theorem statements, and \textsc{PutnamBench} \citep{tsoukalas2024putnambench} extended competition-based evaluation to challenging undergraduate problems.
\textsc{FormalMATH} \citep{yu2025formalmath} expanded the scale and disciplinary coverage of Lean~4 benchmarks, while \textsc{FormalProofBench} \citep{ravi2026formalproofbench} targeted advanced undergraduate and graduate problems from textbooks and qualifying examinations.
Moving toward mathematical practice, \textsc{RLMEval} \citep{poiroux2025rlmeval} evaluates theorems from research-level Lean formalization projects, and \textsc{FormalML} \citep{yang2025formalml} studies subgoal completion in machine-learning theory, including optimization and probability inequalities.

\textbf{Proof automation, representation, and semantic faithfulness.}
Lean~4 \citep{demoura2021lean4} and Mathlib \citep{mathlib2020} provide an extensible proof environment and reusable mathematical abstractions for automated reasoning.
\textsc{LeanDojo} \citep{yang2023leandojo} combines programmatic proof interaction with retrieval-augmented premise selection, while \textsc{Lean Copilot} \citep{song2024leancopilot} integrates tactic suggestion and proof search into interactive formalization.
\textsc{Lean-STaR} \citep{lin2024leanstar} interleaves informal reasoning with tactic generation; \textsc{DeepSeek-Prover-V1.5} \citep{xin2024deepseekprover15} combines proof-assistant feedback with reinforcement learning and tree search; and \textsc{DeepSeek-Prover-V2} \citep{ren2025deepseekprover2} develops reinforcement learning around subgoal decomposition.
These advances address proof construction, but successful checking alone does not establish correspondence with an intended informal claim.
\textsc{FormalAlign} \citep{lu2024formalign} explicitly evaluates informal--formal semantic alignment, while \textsc{MathAtlas} \citep{patel2026mathatlas} examines graduate-level autoformalization with definitions and dependency structure.
\textsc{TaoBench} \citep{taylor2026taobench} isolates a related representation issue through paired, mathematically equivalent statements expressed using bespoke and Mathlib definitions.

\textbf{Formal probability and stochastic-process infrastructure.}
Substantial Lean developments already support the mathematics underlying \stochbench{}.
\citet{ying2022doob} formalize Doob's martingale convergence theorems together with conditional expectation, stopping times, and martingale theory; \citet{marion2025ionescutulcea} constructs trajectory-space probability measures through the Ionescu--Tulcea theorem, and \citet{degenne2025markovkernels} develops Markov kernels and disintegration.
\citet{degenne2025formalizationbrownianmotionlean} formalize Brownian motion and its extension and path-continuity machinery, while \citet{coelho2026ito} develops the $L^2$ It\^o integral and It\^o's formula for $C^3$ functions with bounded derivatives.
Complementary work connects textbook probability to Mathlib interfaces \citep{deng2026probabilitytextbook}, verifies reinforcement-learning convergence \citep{zhang2025rltheory}, and constructs a mathematical-finance library with explicit faithfulness auditing \citep{coelho2026mathfinance}.

\section{The \stochbench{} Benchmark}
\label{sec:benchmark}

\paragraph{Sources and selection.}
\stochbench{} contains 450 Lean~4 theorem targets in graduate stochastic processes. We combine problems written for the benchmark with exercises, lemmas, theorems, and corollaries selected from \emph{Probability, Mathematical Statistics, and Stochastic Processes} \citep{siegrist_prob_stoch} and the MIT course notes and assignments for \emph{Introduction to Stochastic Processes} \citep{wu2015stochastic}, \emph{Advanced Stochastic Processes} \citep{gamarnik2013advanced}, and \emph{Discrete Stochastic Processes} \citep{gallager2011stochastic}. We selected problems for their relevance to stochastic processes and wrote them as claims with hypotheses. Statements that are closer to general probability theory, such as ``show that the total variation distance satisfies triangle inequality : $||\mu - \nu||_{TV} \leq ||\mu - \eta||_{TV} + ||\eta - \nu||_{TV}$'', were not included. The corpus covers eight topics, summarized in Table~\ref{tab:topics}.

\paragraph{Statement construction.}
All benchmark-specific definitions, hypotheses and questions are human-written. An Opus~4.8-based formalizer assisted with expressing the problems as Lean theorem statements. We revised candidate statements using Lean feedback until they elaborated in Lean~4.30.0 with a fixed Mathlib version. Elaboration checks that a statement is well-typed. We consider the task to prove the theorem with established correspondence with the source problem. On the off chance that a formalization error has crept in, we also accept a kernel-checked proof of the theorem being incorrect.

\paragraph{Shared mathematical definitions.}
We build shared abstractions and definitions for recurring concepts. Finite-state chains use a common matrix representation for stochasticity, stationarity, irreducibility, aperiodicity, eventual positivity of transition powers, detailed balance, time reversal, and total-variation distance. \texttt{IsHittingSolution} and \texttt{returnTime} express first-step equations, while \texttt{nstep} defines transition powers through infinite sums for countable-state formulations. Other definitions connect the targets to Mathlib: \texttt{natStop} converts natural-valued stopping times to \texttt{WithTop}, \texttt{runningMax} expresses finite running maxima, and \texttt{IsConstDrift} states conditional increment identities. Reusing these definitions gives related targets a common mathematical representation.

\paragraph{Marginals and joint process laws.}
We distinguish the distribution of a process at one time from its joint behavior over time.
\texttt{HasMatrixMarginals} relates the distribution of $X_n$ to the corresponding row of $P^n$.
\texttt{HasChainLaw} instead specifies finite-dimensional probabilities through
\[
\mathbb{P}_{\mu}(X_0=x_0,\ldots,X_n=x_n)
=
\nu(x_0)\prod_{i=0}^{n-1}P(x_i,x_{i+1}).
\]
The coupling-bound target combines matrix marginals with an explicit condition that the processes agree after the meeting time.
The strong-stationary-time target uses the joint law and stopping-time conditions to relate the state at the stopping time to the state at a later deterministic time.
These representations specify which information about the process is available to the prover.

\textbf{Formalization scope.}
Some problems require infrastructure unavailable in the Mathlib environment. \emph{Direct} targets use Mathlib objects or shared definitions, while \emph{abstracted} targets take the required properties as hypotheses. The JSON records these labels as \texttt{literal} and \texttt{abstract}, respectively. The supplied properties may define an object or provide intermediate results from the source problem. These are different choices: specifying Brownian-motion properties does not assume a quadratic-variation conclusion, whereas assuming memorylessness removes the need to derive it from continuous-time chain dynamics. Likewise, a hitting-time target stated through first-step equations need not establish that their solution equals a pathwise expected hitting time. 

\textbf{Path properties and convergence.}
The targets state the required form of convergence explicitly.
Brownian-motion properties are expressed through Gaussian increment laws, independence, and almost-sure path continuity; several targets package these properties in a local \texttt{IsBM} definition.
The quadratic-variation target asks for convergence of the mean-square error as the partition mesh tends to zero.
The Donsker target asks for convergence of expectations for every bounded continuous functional on $C([0,T],\mathbb{R})$, rather than only convergence at individual times.
A separate target asks for existence and uniqueness of Wiener measure on continuous path space.

\textbf{Human review and release.}
We reviewed the definitions and hypotheses against the source problems, but they would benefit from further peer review. Lean verifies that each completed proof establishes its conclusion under the stated hypotheses; source review assesses whether the definitions and hypotheses represent the intended problem. Each JSON record contains an identifier, a problem name, an informal statement, a Lean target, and a representation label.
We also release the shared definitions and baseline proof attempts. The release is a collection of theorem targets, not a claim that all targets have complete proofs. Baseline proof checking and results are described in Section~\ref{sec:evaluation}.

\begin{table}[t]
\centering
\captionsetup{position=top,skip=7pt}
\captionsetup[subtable]{hypcap=false}
\captionsetup{subrefformat=simple}
\renewcommand{\thesubtable}{(\alph{subtable})}

\label{tab:corpus-results}

\small
\setlength{\tabcolsep}{3pt}
\setlength{\heavyrulewidth}{0.6pt}
\setlength{\lightrulewidth}{0.35pt}
\setlength{\cmidrulewidth}{0.35pt}
\renewcommand{\arraystretch}{1.10}
\sisetup{
    group-digits=false,
    table-number-alignment=right,
    table-text-alignment=right
}

\begin{tabular*}{\linewidth}{
    @{\extracolsep{\fill}}l
    S[table-format=3.0]
    S[table-format=3.0]
    S[table-format=3.0]
    @{\hspace{1em}}
    S[table-format=2.0]
    S[table-format=2.0]
    S[table-format=3.0]
    S[table-format=2.1]@{}
}
    \toprule
    & \multicolumn{3}{c}{%
        \phantomsubcaption\phantomsection\label{tab:topics}%
        \textit{\thesubtable\ Corpus composition}}
    & \multicolumn{4}{c}{%
        \phantomsubcaption\phantomsection\label{tab:bytopic}%
        \textit{\thesubtable\ Baseline}} \\
    \cmidrule(lr){2-4}\cmidrule(l){5-8}
    & & & & \multicolumn{3}{c}{Clean proofs} & \\
    \cmidrule(lr){5-7}
    Topic
        & {Direct} & {Abstracted} & {Items}
        & {Direct} & {Abstracted} & {Total} & {Rate (\%)} \\
    \midrule

    Poisson processes
        & 5 & 35 & 40
        & 0 & 7 & 7 & 17.5 \\
    Markov chains (finite \& countable)
        & 23 & 73 & 96
        & 15 & 22 & 37 & 38.5 \\
    Renewal processes
        & 0 & 41 & 41
        & 0 & 2 & 2 & 4.9 \\
    Continuous-time Markov \& queues
        & 1 & 54 & 55
        & 1 & 22 & 23 & 41.8 \\
    Random walks \& large deviations
        & 9 & 53 & 62
        & 6 & 10 & 16 & 25.8 \\
    Martingales \& stopping
        & 67 & 27 & 94
        & 51 & 7 & 58 & 61.7 \\
    Brownian motion \& stochastic calculus
        & 0 & 45 & 45
        & 0 & 8 & 8 & 17.8 \\
    Weak convergence \& functional limits
        & 9 & 8 & 17
        & 6 & 0 & 6 & 35.3 \\
    \midrule
    \textit{All}
        & 114 & 336 & 450
        & 79 & 78 & 157 & 34.9 \\

    \addlinespace[1.1ex]
    \multicolumn{4}{@{}l}{%
        \phantomsubcaption\phantomsection\label{tab:representation}%
        \textit{\thesubtable\ Clean-proof rates by target class (\%)}}
        & \multicolumn{1}{r}{69.3}
        & \multicolumn{1}{r}{23.2}
        & & \\
    \bottomrule
\end{tabular*}

\par\vspace{4pt}
\begin{minipage}{\linewidth}
\end{minipage}
\caption{Corpus composition and baseline results for \stochbench{}
(recorded proving time at most 15 minutes). Only proofs with recorded proving times of at most 900 seconds are counted. Topic rates use all items in the topic; class rates use all items in the class. The class comparison is descriptive, not a controlled causal effect.}
\end{table}

\section{Evaluation}
\label{sec:evaluation}

We evaluated a multi-turn tool-using Opus~4.8-based agent using \texttt{lean4skills} and the Lean LSP MCP server \citep{lean4-skills, lean-lsp-mcp}. Each target received one run capped at 15 minutes, allowing Lean-error inspection, library and shared-definition search, \texttt{loogle} and \texttt{leansearch} queries, and proof revisions. The same model family assisted with statement construction. This is a single-agent, single-budget baseline, not a model comparison or repeated-run evaluation.

The agent produced 157 clean proofs out of 450. A proof is \emph{clean} if Lean accepts it without \texttt{sorry}, \texttt{sorryAx}, or additional admitted facts, checked by Lean comparator. Tables~\ref{tab:bytopic} and~\ref{tab:representation} report topic and class breakdowns. These are descriptive comparisons: they do not separate abstraction effects from differences in problems or library support.

Qualitative inspection found proof-search failures on plausible targets, missing lemmas or difficult library interfaces, and a smaller group of formalization defects, including missing measurability, integrability, or non-emptiness assumptions.

\section{Limitations and Conclusion}
\label{sec:conclusion}

We note that \stochbench{}'s question curation, faithfulness review, and its topic and direct/abstracted classifications are currently decided by human curators, introducing some bias; as the corresponding terminologies were not rigorously defined within the scope of this work. Despite these limitations, \stochbench{} provides a focused testbed for evaluating proof agents on graduate stochastic processes.
Its newly constructed informal--formal pairs can support autoformalization training, while successfully checked baseline proofs provide supervision for proof generation.
Together with the shared definitions, these resources support both the development of stronger domain-specific provers and the continued formalization of stochastic processes in Lean.

\bibliographystyle{plainnat}
\bibliography{custom}

@inproceedings{zheng2022minif2f,
  title = {{miniF2F}: A Cross-System Benchmark for Formal Olympiad-Level Mathematics},
  author = {Zheng, Kunhao and Han, Jesse Michael and Polu, Stanislas},
  booktitle = {International Conference on Learning Representations},
  year = {2021},
  journal = {International Conference on Learning Representations},
}

@article{azerbayev2023proofnet,
  title = {{ProofNet}: Autoformalizing and Formally Proving Undergraduate-Level Mathematics},
  author = {Azerbayev, Zhangir and Piotrowski, Bartosz and Schoelkopf, Hailey and Ayers, Edward W. and Radev, Dragomir and Avigad, Jeremy},
  journal = {arXiv.org},
  year = {2023},
  doi = {10.48550/arXiv.2302.12433},
}

@article{tsoukalas2024putnambench,
  title = {{PutnamBench}: Evaluating Neural Theorem-Provers on the Putnam Mathematical Competition},
  author = {Tsoukalas, George and Lee, Jasper and Jennings, John and Xin, Jimmy and Ding, Michelle and Jennings, Michael and Thakur, Amitayush and Chaudhuri, Swarat},
  journal = {Advances in Neural Information Processing Systems 37},
  year = {2024},
  pages = {11545-11569},
  doi = {10.52202/079017-0368},
  publisher = {Neural Information Processing Systems Foundation, Inc. (NeurIPS)},
}

@article{ravi2026formalproofbench,
  title = {{FormalProofBench}: Can Models Write Graduate Level Math Proofs That Are Formally Verified?},
  author = {Ravi, Nikil and Ying, Kexing and Nesterov, Vasilii and Krishnan, Rayan and Uskuplu, Elif and Xia, Bingyu and Aswedige, Janitha and Nashold, Langston},
  journal = {arXiv.org},
  year = {2026},
  doi = {10.48550/arXiv.2603.26996},
}

@inproceedings{demoura2021lean4,
  title = {The {Lean 4} Theorem Prover and Programming Language},
  author = {Moura, L. D. and Ullrich, Sebastian},
  booktitle = {CADE},
  year = {2021},
  journal = {CADE},
  pages = {625-635},
  doi = {10.1007/978-3-030-79876-5_37},
  publisher = {Springer International Publishing},
}

@software{lean-lsp-mcp,
  author = {Oliver Dressler},
  title = {{Lean LSP MCP: Tools for agentic interaction with the Lean theorem prover}},
  url = {https://github.com/oOo0oOo/lean-lsp-mcp},
  month = {3},
  year = {2025}
}

@software{lean4-skills,
  author = {Cameron Freer},
  title = {Lean 4 {Skills}: Theorem proving skill and workflow pack for {AI} coding agents},
  url = {https://github.com/cameronfreer/lean4-skills},
  month = oct,
  year = {2025}
}

@misc{hariharan2026milestoneformalizationspherepacking,
      title={A Milestone in Formalization: The Sphere Packing Problem in Dimension 8}, 
      author={Sidharth Hariharan and Christopher Birkbeck and Seewoo Lee and Ho Kiu Gareth Ma and Bhavik Mehta and Auguste Poiroux and Maryna Viazovska},
      year={2026},
      eprint={2604.23468},
      archivePrefix={arXiv},
      primaryClass={math.MG},
      url={https://arxiv.org/abs/2604.23468}, 
}

@misc{wu2015stochastic,
  author       = {Hao Wu},
  year = {2015},
  title        = {18.445 {Introduction} to {Stochastic} {Processes}},
  howpublished = {Spring 2015. Massachusetts Institute of Technology: MIT OpenCourseWare, \url{https://ocw.mit.edu/}},
  note         = {License: Creative Commons BY-NC-SA}
}

@misc{gamarnik2013advanced,
  author       = {David Gamarnik},
  year         = {2013},
  title        = {15.070{J} {Advanced} {Stochastic} {Processes}},
  howpublished = {Fall 2013. Massachusetts Institute of Technology: MIT OpenCourseWare, \url{https://ocw.mit.edu/}},
  note         = {License: Creative Commons BY-NC-SA}
}

@misc{gallager2011stochastic,
  author       = {Robert Gallager},
  year = {2011},
  title        = {6.262 {Discrete} {Stochastic} {Processes}},
  howpublished = {Spring 2011. Massachusetts Institute of Technology: MIT OpenCourseWare, \url{https://ocw.mit.edu/}},
  note         = {License: Creative Commons BY-NC-SA}
}

@misc{siegrist_prob_stoch,
  author       = {Kyle Siegrist},
  title        = {Probability, {Mathematical} {Statistics}, and {Stochastic} {Processes}},
  year         = {2022},
  howpublished = {LibreTexts},
  url          = {https://stats.libretexts.org/Bookshelves/Probability_Theory/Probability_Mathematical_Statistics_and_Stochastic_Processes_(Siegrist)},
  note         = {Originally sourced from \url{http://www.randomservices.org/random}. License: CC BY 2.0}
}

@misc{best2025completeformalizationfermatstheorem,
      title={A complete formalization of Fermat's Last Theorem for regular primes in Lean}, 
      author={Alex Best and Christopher Birkbeck and Riccardo Brasca and Eric Rodriguez Boidi and Ruben van De Velde and Andrew Yang},
      year={2025},
      eprint={2410.01466},
      archivePrefix={arXiv},
      primaryClass={cs.FL},
      url={https://arxiv.org/abs/2410.01466}, 
}

@misc{degenne2025formalizationbrownianmotionlean,
      title={Formalization of Brownian motion in Lean}, 
      author={Rémy Degenne and David Ledvinka and Etienne Marion and Peter Pfaffelhuber},
      year={2025},
      eprint={2511.20118},
      archivePrefix={arXiv},
      primaryClass={math.PR},
      url={https://arxiv.org/abs/2511.20118}, 
}

@misc{undergrad_todo,
  author       = {{The mathlib Community}},
  title        = {Missing undergraduate mathematics in mathlib},
  howpublished = {\url{https://leanprover-community.github.io/undergrad_todo.html}},
  year         = {2026},
}

@misc{yu2025formalmath,
  title = {{FormalMATH: Benchmarking Formal Mathematical Reasoning of Large Language Models}},
  author = {Zhouliang Yu and Ruotian Peng and Keyi Ding and Yizhe Li and Zhongyuan Peng and Minghao Liu and Yifan Zhang and Zheng Yuan and Huajian Xin and Wenhao Huang and Yandong Wen and Ge Zhang and Weiyang Liu},
  year = {2025},
  eprint = {2505.02735},
  archivePrefix = {arXiv},
  doi = {10.48550/arXiv.2505.02735},
  url = {https://arxiv.org/abs/2505.02735},
}

@misc{poiroux2025rlmeval,
  title = {{RLMEval: Evaluating Research-Level Neural Theorem Proving}},
  author = {Auguste Poiroux and Antoine Bosselut and Viktor Kun{\v c}ak},
  year = {2025},
  eprint = {2510.25427},
  archivePrefix = {arXiv},
  doi = {10.48550/arXiv.2510.25427},
  url = {https://arxiv.org/abs/2510.25427},
}

@misc{yang2025formalml,
  title = {{FormalML: A Benchmark for Evaluating Formal Subgoal Completion in Machine Learning Theory}},
  author = {Xiao-Wen Yang and Zihao Zhang and Jianuo Cao and Zhi Zhou and Zenan Li and Lan-Zhe Guo and Yuan Yao and Taolue Chen and Yu-Feng Li and Xiaoxing Ma},
  year = {2025},
  eprint = {2510.02335},
  archivePrefix = {arXiv},
  doi = {10.48550/arXiv.2510.02335},
  url = {https://arxiv.org/abs/2510.02335},
}

@inproceedings{mathlib2020,
  title = {{The Lean Mathematical Library}},
  author = {{The mathlib Community}},
  year = {2020},
  booktitle = {Proceedings of the 9th ACM SIGPLAN International Conference on Certified Programs and Proofs},
  pages = {367--381},
  publisher = {ACM},
  doi = {10.1145/3372885.3373824},
  url = {https://doi.org/10.1145/3372885.3373824},
}

@misc{yang2023leandojo,
  title = {{LeanDojo: Theorem Proving with Retrieval-Augmented Language Models}},
  author = {Kaiyu Yang and Aidan M. Swope and Alex Gu and Rahul Chalamala and Peiyang Song and Shixing Yu and Saad Godil and Ryan Prenger and Anima Anandkumar},
  year = {2023},
  eprint = {2306.15626},
  archivePrefix = {arXiv},
  doi = {10.48550/arXiv.2306.15626},
  url = {https://arxiv.org/abs/2306.15626},
}

@misc{song2024leancopilot,
  title = {{Towards Large Language Models as Copilots for Theorem Proving in Lean}},
  author = {Peiyang Song and Kaiyu Yang and Anima Anandkumar},
  year = {2024},
  eprint = {2404.12534},
  archivePrefix = {arXiv},
  doi = {10.48550/arXiv.2404.12534},
  url = {https://arxiv.org/abs/2404.12534},
}

@misc{lin2024leanstar,
  title = {{Lean-STaR: Learning to Interleave Thinking and Proving}},
  author = {Haohan Lin and Zhiqing Sun and Yiming Yang and Sean Welleck},
  year = {2024},
  eprint = {2407.10040},
  archivePrefix = {arXiv},
  doi = {10.48550/arXiv.2407.10040},
  url = {https://arxiv.org/abs/2407.10040},
}

@misc{xin2024deepseekprover15,
  title = {{DeepSeek-Prover-V1.5: Harnessing Proof Assistant Feedback for Reinforcement Learning and Monte-Carlo Tree Search}},
  author = {Huajian Xin and Z. Z. Ren and Junxiao Song and Zhihong Shao and Wanjia Zhao and Haocheng Wang and Bo Liu and Liyue Zhang and Xuan Lu and Qiushi Du and Wenjun Gao and Qihao Zhu and Dejian Yang and Zhibin Gou and Z. F. Wu and Fuli Luo and Chong Ruan},
  year = {2024},
  eprint = {2408.08152},
  archivePrefix = {arXiv},
  doi = {10.48550/arXiv.2408.08152},
  url = {https://arxiv.org/abs/2408.08152},
}

@misc{ren2025deepseekprover2,
  title = {{DeepSeek-Prover-V2: Advancing Formal Mathematical Reasoning via Reinforcement Learning for Subgoal Decomposition}},
  author = {Z. Z. Ren and Zhihong Shao and Junxiao Song and Huajian Xin and Haocheng Wang and Wanjia Zhao and Liyue Zhang and Zhe Fu and Qihao Zhu and Dejian Yang and Z. F. Wu and Zhibin Gou and Shirong Ma and Hongxuan Tang and Yuxuan Liu and Wenjun Gao and Daya Guo and Chong Ruan},
  year = {2025},
  eprint = {2504.21801},
  archivePrefix = {arXiv},
  doi = {10.48550/arXiv.2504.21801},
  url = {https://arxiv.org/abs/2504.21801},
}

@misc{lu2024formalign,
  title = {{FormalAlign: Automated Alignment Evaluation for Autoformalization}},
  author = {Jianqiao Lu and Yingjia Wan and Yinya Huang and Jing Xiong and Zhengying Liu and Zhijiang Guo},
  year = {2024},
  eprint = {2410.10135},
  archivePrefix = {arXiv},
  doi = {10.48550/arXiv.2410.10135},
  url = {https://arxiv.org/abs/2410.10135},
}

@misc{patel2026mathatlas,
  title = {{MathAtlas: A Benchmark for Autoformalization in the Wild}},
  author = {Nilay Patel and Noah Arias and Davit Babayan and Victoria Cochran and Timothy Libman and Hafsah Mahmood and Liam McCarty and Soli Munoz and Laurel Willey and Jeffrey Flanigan},
  year = {2026},
  eprint = {2605.14061},
  archivePrefix = {arXiv},
  doi = {10.48550/arXiv.2605.14061},
  url = {https://arxiv.org/abs/2605.14061},
}

@misc{taylor2026taobench,
  title = {{TaoBench: Do Automated Theorem Prover LLMs Generalize Beyond MathLib?}},
  author = {Alexander K Taylor and Junyi Zhang and Ethan Ji and Vigyan Sahai and Haikang Deng and Yuanzhou Chen and Yifan Yuan and Di Wu and Jia-Chen Gu and Kai-Wei Chang and Nanyun Peng and Amit Sahai and Wei Wang},
  year = {2026},
  eprint = {2603.12744},
  archivePrefix = {arXiv},
  doi = {10.48550/arXiv.2603.12744},
  url = {https://arxiv.org/abs/2603.12744},
}

@misc{ying2022doob,
  title = {{A Formalization of Doob's Martingale Convergence Theorems in mathlib}},
  author = {Kexing Ying and R{\'e}my Degenne},
  year = {2022},
  eprint = {2212.05578},
  archivePrefix = {arXiv},
  doi = {10.48550/arXiv.2212.05578},
  url = {https://arxiv.org/abs/2212.05578},
}

@misc{marion2025ionescutulcea,
  title = {{A Formalization of the Ionescu-Tulcea Theorem in Mathlib}},
  author = {Etienne Marion},
  year = {2025},
  eprint = {2506.18616},
  archivePrefix = {arXiv},
  doi = {10.48550/arXiv.2506.18616},
  url = {https://arxiv.org/abs/2506.18616},
}

@misc{degenne2025markovkernels,
  title = {{Markov Kernels in Mathlib's Probability Library}},
  author = {R{\'e}my Degenne},
  year = {2025},
  eprint = {2510.04070},
  archivePrefix = {arXiv},
  doi = {10.48550/arXiv.2510.04070},
  url = {https://arxiv.org/abs/2510.04070},
}

@misc{coelho2026ito,
  title = {{A Machine-Checked It{\^o} Calculus for Brownian Motion}},
  author = {Raphael Coelho},
  year = {2026},
  eprint = {2606.15089},
  archivePrefix = {arXiv},
  doi = {10.48550/arXiv.2606.15089},
  url = {https://arxiv.org/abs/2606.15089},
}

@misc{deng2026probabilitytextbook,
  title = {{From Lecture Notes to Lean: Formalizing a Textbook on Probability Theory}},
  author = {Shuo Deng and Kenneth W. Shum},
  year = {2026},
  eprint = {2607.27298},
  archivePrefix = {arXiv},
  doi = {10.48550/arXiv.2607.27298},
  url = {https://arxiv.org/abs/2607.27298},
}

@misc{zhang2025rltheory,
  title = {{Towards Formalizing Reinforcement Learning Theory: A Robbins-Siegmund Approach}},
  author = {Shangtong Zhang},
  year = {2025},
  eprint = {2511.03618},
  archivePrefix = {arXiv},
  doi = {10.48550/arXiv.2511.03618},
  url = {https://arxiv.org/abs/2511.03618},
}

@misc{coelho2026mathfinance,
  title = {{A Formally Verified Library of Mathematical Finance in Lean 4}},
  author = {Raphael Coelho},
  year = {2026},
  eprint = {2606.01356},
  archivePrefix = {arXiv},
  doi = {10.48550/arXiv.2606.01356},
  url = {https://arxiv.org/abs/2606.01356},
}

\newpage

\appendix
\section{Supplementary material}
\subsection{An abstracted proof example - SOTA Prover}
\label{sec:q361-abstracted-proof}


We illustrate Q361 through its natural-language statement, Lean abstraction, and an agent-generated proof obtained in a separate run lasting more than 30 minutes, outside the 15-minute baseline protocol.

\paragraph{Natural-language statement.}
Suppose that a Markov chain on a finite, nonempty state space $\Omega$ is
irreducible and has stationary probability measure $\pi$. Define the hitting
time of $x \in \Omega$ by
\[
  \tau_x = \min\{n \geq 0 : X_n = x\}.
\]
For a fixed state $a \in \Omega$, let
\[
  t_{\odot} = \sum_{x \in \Omega} \mathbb{E}_a[\tau_x]\,\pi(x),
  \qquad
  t_{\mathrm{hit}} = \max_{x,y \in \Omega}\mathbb{E}_x[\tau_y]
  \geq t_{\odot}.
\]
Show that
\[
  t_{\mathrm{hit}}
  \leq 2\max_{w \in \Omega}\mathbb{E}_{\pi}[\tau_w],
  \qquad
  \mathbb{E}_{\pi}[\tau_w]
  = \sum_{x \in \Omega}\pi(x)\mathbb{E}_x[\tau_w].
\]

\paragraph{Formalization.}
The Lean statement represents expected hitting times by a real-valued function
$g(x,y)$ satisfying the first-step equations
\[
  g(x,y) =
  \begin{cases}
    0, & x=y,\\
    1 + \sum_{z \in \Omega} P(x,z)g(z,y), & x\neq y.
  \end{cases}
\]
These equations are supplied by \texttt{IsHittingSolution}; the proof works
with this characterization rather than constructing hitting-time random
variables. The assumptions \texttt{IsStochastic}, \texttt{IsIrreducible}, and
\texttt{IsStationary} specify the transition matrix and stationary distribution.
The auxiliary quantity $t_{\odot}$ and the given lower bound on
$t_{\mathrm{hit}}$ are not needed for the formalized conclusion.


\paragraph{Proof structure and difficulty.}
Although Q361 asks for a single inequality, the generated proof develops nine auxiliary theorems across several levels of abstraction. It establishes nonnegativity of matrix powers and hitting-time solutions, extends closure under positive one-step transitions to positive matrix powers, and proves a maximum-principle propagation lemma. A return-time identity yields harmonicity of Kemeny's function $K(x)=\sum_y\pi(y)g(x,y)$, whose constancy follows from irreducibility and the maximum principle. The same propagation lemmas are reused for $\theta(x)=g(x,b)-g(x,c)$ to establish the hitting-time triangle inequality. Stationary averaging then combines these results into a pointwise bound, which is lifted to the required supremum inequality.

\paragraph{Implications for proof search.}
The abstraction supplies the first-step equations, not Kemeny constancy or the triangle inequality. The proof therefore builds supporting mathematics rather than merely simplifying the supplied hypotheses. Its structure illustrates a challenge for hierarchical proof search: identifying useful intermediate statements, proving them at sufficient generality for reuse, and coordinating their dependencies. Here, elementary matrix and finite-sum reasoning supports harmonic-function arguments, which in turn support the global hitting-time bound. The agent-produced result is thus a coherent collection of auxiliary theorems together with their final composition. This example shows how an abstracted benchmark target can test mathematical proof construction beyond local tactic selection.


\begin{lstlisting}
import Mathlib
import Leanproj.Basic
open MeasureTheory ProbabilityTheory Filter Auto
open scoped ENNReal NNReal Topology Matrix RealInnerProductSpace
attribute [local instance] Classical.propDecidable
set_option maxHeartbeats 2000000
set_option maxRecDepth 4096

namespace Q361Helpers

open Finset

/-- Entries of powers of a nonnegative matrix are nonnegative. -/
theorem pow_nonneg_entry {Ω : Type*} [Fintype Ω] [DecidableEq Ω]
    (P : Matrix Ω Ω ℝ) (hP1 : ∀ x y, 0 ≤ P x y) : ∀ n x y, 0 ≤ (P ^ n) x y := by
  intro n
  induction n with
  | zero => intro x y; simp [pow_zero, Matrix.one_apply]; split <;> norm_num
  | succ k ih =>
    intro x y
    rw [pow_succ, Matrix.mul_apply]
    apply Finset.sum_nonneg; intro z _
    exact mul_nonneg (ih x z) (hP1 z y)

/-- Hitting times are nonnegative. -/
theorem g_nonneg {Ω : Type*} [Fintype Ω] [Nonempty Ω] [DecidableEq Ω]
    (P : Matrix Ω Ω ℝ) (hP1 : ∀ x y, 0 ≤ P x y) (hP2 : ∀ x, ∑ y, P x y = 1)
    (g : Ω → Ω → ℝ) (hg : ∀ x y, g x y = if x = y then 0 else 1 + ∑ z, P x z * g z y)
    (y : Ω) : ∀ x, 0 ≤ g x y := by 
    obtain ⟨x0, -, hx0⟩ := Finset.exists_min_image Finset.univ (fun x => g x y)
    ⟨Classical.arbitrary Ω, Finset.mem_univ _⟩
  have hmin : ∀ x, g x0 y ≤ g x y := fun x => hx0 x (Finset.mem_univ x)
  have hx0y : g x0 y = 0 := by
    by_cases h : x0 = y
    · rw [hg x0 y, if_pos h]
    · exfalso
      have hval := hg x0 y
      rw [if_neg h] at hval
      have hlb : (1 : ℝ) + ∑ z, P x0 z * g z y ≥ 1 + ∑ z, P x0 z * g x0 y := by
        have : (∑ z, P x0 z * g z y) ≥ ∑ z, P x0 z * g x0 y := by
          apply Finset.sum_le_sum; intro z _
          exact mul_le_mul_of_nonneg_left (hmin z) (hP1 x0 z)
        linarith
      have hsum : (∑ z, P x0 z * g x0 y) = g x0 y := by
        rw [← Finset.sum_mul, hP2 x0, one_mul]
      rw [hsum] at hlb
      linarith [hval, hlb]
  intro x
  calc 0 = g x0 y := hx0y.symm
    _ ≤ g x y := hmin x

/-- Forward-closure reachability: a set closed under positive one-step transitions is
closed under positive `n`-step transitions. -/
theorem reach_closed {Ω : Type*} [Fintype Ω] [DecidableEq Ω]
    (P : Matrix Ω Ω ℝ) (hP1 : ∀ x y, 0 ≤ P x y)
    (A : Ω → Prop) (hclosed : ∀ x, A x → ∀ z, 0 < P x z → A z)
    (x0 : Ω) (hx0 : A x0) : ∀ n z, 0 < (P ^ n) x0 z → A z := by
  have hpow := pow_nonneg_entry P hP1
  intro n
  induction n with
  | zero =>
    intro z hz
    rw [pow_zero, Matrix.one_apply] at hz
    by_cases h : x0 = z
    · rwa [← h]
    · simp [h] at hz
  | succ k ih =>
    intro z hz
    rw [pow_succ, Matrix.mul_apply] at hz
    have hex : ∃ w ∈ Finset.univ, (0:ℝ) < (P ^ k) x0 w * P w z := by
      apply Finset.exists_lt_of_sum_lt; simpa using hz
    obtain ⟨w, -, hw⟩ := hex
    have h1 : 0 < (P ^ k) x0 w :=
      lt_of_le_of_ne (hpow k x0 w) (fun h => by rw [← h, zero_mul] at hw; exact lt_irrefl _ hw)
    have h2 : 0 < P w z :=
      lt_of_le_of_ne (hP1 w z) (fun h => by rw [← h, mul_zero] at hw; exact lt_irrefl _ hw)
    exact hclosed w (ih w h1) z h2

/-- Maximum-principle propagation step: at a maximizing point where `f` is harmonic, all
positively-reachable neighbours also attain the maximum. -/
theorem prop_step {Ω : Type*} [Fintype Ω]
    (P : Matrix Ω Ω ℝ) (hP1 : ∀ x y, 0 ≤ P x y) (hP2 : ∀ x, ∑ y, P x y = 1)
    (f : Ω → ℝ) (m : ℝ) (hmax : ∀ z, f z ≤ m) (x : Ω) (hfx : f x = m)
    (hharm : f x = ∑ z, P x z * f z) (z : Ω) (hz : 0 < P x z) : f z = m := by
  have hsum0 : (∑ w, P x w * (m - f w)) = 0 := by
    have : (∑ w, P x w * (m - f w)) = (∑ w, P x w) * m - ∑ w, P x w * f w := by
      rw [Finset.sum_mul, ← Finset.sum_sub_distrib]
      apply Finset.sum_congr rfl; intro w _; ring
    rw [this, hP2 x, one_mul, ← hharm, hfx]; ring
  have hnonneg : ∀ w ∈ Finset.univ, 0 ≤ P x w * (m - f w) := by
    intro w _; exact mul_nonneg (hP1 x w) (by linarith [hmax w])
  have hzero := (Finset.sum_eq_zero_iff_of_nonneg hnonneg).mp hsum0 z (Finset.mem_univ z)
  have : m - f z = 0 := by
    rcases mul_eq_zero.mp hzero with h | h
    · linarith [hz]
    · exact h
  linarith

/-- Return-time identity: `π w · (1 + ∑_z P(w,z) g(z,w)) = 1`. -/
theorem returnTime_identity {Ω : Type*} [Fintype Ω] [Nonempty Ω] [DecidableEq Ω]
    (P : Matrix Ω Ω ℝ) (g : Ω → Ω → ℝ)
    (hg : ∀ x y, g x y = if x = y then 0 else 1 + ∑ z, P x z * g z y)
    (π : Ω → ℝ) (hpi2 : ∑ x, π x = 1)
    (hpi3 : ∀ y, ∑ x, π x * P x y = π y) (w : Ω) :
    π w * (1 + ∑ z, P w z * g z w) = 1 := by
  have hswap : (∑ x, π x * ∑ z, P x z * g z w) = ∑ x, π x * g x w := by
    have e1 : (∑ x, π x * ∑ z, P x z * g z w) = ∑ x, ∑ z, π x * (P x z * g z w) := by
      apply Finset.sum_congr rfl; intro x _; rw [Finset.mul_sum]
    rw [e1, Finset.sum_comm]
    apply Finset.sum_congr rfl; intro z _
    have : (∑ x, π x * (P x z * g z w)) = (∑ x, π x * P x z) * g z w := by
      rw [Finset.sum_mul]; apply Finset.sum_congr rfl; intro x _; ring
    rw [this, hpi3 z]
  have s1 : (∑ x, if x = w then 0 else π x) = 1 - π w := by
    have : (∑ x, if x = w then 0 else π x) = ∑ x, (π x - if x = w then π x else 0) := by
      apply Finset.sum_congr rfl; intro x _; by_cases hx : x = w <;> simp [hx]
    rw [this, Finset.sum_sub_distrib, hpi2, Finset.sum_ite_eq' Finset.univ w π]; simp
  have s2 : (∑ x, if x = w then 0 else π x * ∑ z, P x z * g z w)
      = (∑ x, π x * g x w) - π w * ∑ z, P w z * g z w := by
    have step1 : (∑ x, if x = w then 0 else π x * ∑ z, P x z * g z w)
        = (∑ x, π x * ∑ z, P x z * g z w) - π w * ∑ z, P w z * g z w := by
      have : (∑ x, if x = w then 0 else π x * ∑ z, P x z * g z w)
          = ∑ x, ((π x * ∑ z, P x z * g z w)
              - (if x = w then π w * ∑ z, P w z * g z w else 0)) := by
        apply Finset.sum_congr rfl; intro x _; by_cases hx : x = w <;> simp [hx]
      rw [this, Finset.sum_sub_distrib, Finset.sum_ite_eq' Finset.univ w]; simp
    rw [step1, hswap]
  have key : (∑ x, π x * g x w)
      = (1 - π w) + ((∑ x, π x * g x w) - π w * (∑ z, P w z * g z w)) := by
    have e1 : (∑ x, π x * g x w)
        = ∑ x, π x * (if x = w then 0 else 1 + ∑ z, P x z * g z w) := by
      apply Finset.sum_congr rfl; intro x _; rw [hg x w]
    have e2 : (∑ x, π x * (if x = w then 0 else 1 + ∑ z, P x z * g z w))
        = (∑ x, if x = w then 0 else π x)
          + (∑ x, if x = w then 0 else π x * ∑ z, P x z * g z w) := by
      rw [← Finset.sum_add_distrib]
      apply Finset.sum_congr rfl; intro x _
      by_cases hx : x = w <;> simp [hx] <;> ring
    conv_lhs => rw [e1, e2, s1, s2]
  nlinarith [key]

/-- Kemeny function `K x = ∑_c π_c g(x,c)` is harmonic: `K x = ∑_z P(x,z) K(z)`. -/
theorem K_harmonic {Ω : Type*} [Fintype Ω] [Nonempty Ω] [DecidableEq Ω]
    (P : Matrix Ω Ω ℝ) (hP1 : ∀ x y, 0 ≤ P x y) (hP2 : ∀ x, ∑ y, P x y = 1)
    (g : Ω → Ω → ℝ) (hg : ∀ x y, g x y = if x = y then 0 else 1 + ∑ z, P x z * g z y)
    (π : Ω → ℝ) (hpi1 : ∀ x, 0 ≤ π x) (hpi2 : ∑ x, π x = 1)
    (hRT : ∀ w, π w * (1 + ∑ z, P w z * g z w) = 1) (x : Ω) :
    (∑ c, π c * g x c) = ∑ z, P x z * (∑ c, π c * g z c) := by
  have hpx : π x * (1 + ∑ z, P x z * g z x) = 1 := hRT x
  have hpxpos : 0 < π x := by
    rcases (hpi1 x).lt_or_eq with h | h
    · exact h
    · exfalso; rw [← h, zero_mul] at hpx; norm_num at hpx
  have hret : π x * (∑ z, P x z * g z x) = 1 - π x := by nlinarith [hpx]
  have e1 : (∑ c, π c * g x c) = ∑ c, (if c = x then 0 else π c * (1 + ∑ z, P x z * g z c)) := by
    apply Finset.sum_congr rfl; intro c _
    rw [hg x c]; by_cases h : x = c
    · simp [h]
    · rw [if_neg h, if_neg (Ne.symm h)]
  have e2 : (∑ c, (if c = x then 0 else π c * (1 + ∑ z, P x z * g z c)))
      = (∑ c, if c = x then 0 else π c)
        + (∑ c, if c = x then 0 else π c * ∑ z, P x z * g z c) := by
    rw [← Finset.sum_add_distrib]; apply Finset.sum_congr rfl; intro c _
    by_cases h : c = x <;> simp [h] <;> ring
  have sa : (∑ c, if c = x then 0 else π c) = 1 - π x := by
    have : (∑ c, if c = x then 0 else π c) = ∑ c, (π c - if c = x then π c else 0) := by
      apply Finset.sum_congr rfl; intro c _; by_cases h : c = x <;> simp [h]
    rw [this, Finset.sum_sub_distrib, hpi2, Finset.sum_ite_eq' Finset.univ x π]; simp
  have sb : (∑ c, if c = x then 0 else π c * ∑ z, P x z * g z c)
      = (∑ z, P x z * (∑ c, π c * g z c)) - π x * (∑ z, P x z * g z x) := by
    have drop : (∑ c, if c = x then 0 else π c * ∑ z, P x z * g z c)
        = (∑ c, π c * ∑ z, P x z * g z c) - π x * ∑ z, P x z * g z x := by
      have : (∑ c, if c = x then 0 else π c * ∑ z, P x z * g z c)
          = ∑ c, ((π c * ∑ z, P x z * g z c)
              - (if c = x then π x * ∑ z, P x z * g z x else 0)) := by
        apply Finset.sum_congr rfl; intro c _; by_cases h : c = x <;> simp [h]
      rw [this, Finset.sum_sub_distrib, Finset.sum_ite_eq' Finset.univ x]; simp
    rw [drop]
    congr 1
    have l1 : (∑ c, π c * ∑ z, P x z * g z c) = ∑ c, ∑ z, π c * (P x z * g z c) := by
      apply Finset.sum_congr rfl; intro c _; rw [Finset.mul_sum]
    have l2 : (∑ z, P x z * (∑ c, π c * g z c)) = ∑ z, ∑ c, P x z * (π c * g z c) := by
      apply Finset.sum_congr rfl; intro z _; rw [Finset.mul_sum]
    rw [l1, l2, Finset.sum_comm]
    apply Finset.sum_congr rfl; intro z _; apply Finset.sum_congr rfl; intro c _; ring
  rw [e1, e2, sa, sb, hret]
  ring

/-- The Kemeny function is constant on an irreducible chain. -/
theorem K_const {Ω : Type*} [Fintype Ω] [Nonempty Ω] [DecidableEq Ω]
    (P : Matrix Ω Ω ℝ) (hP1 : ∀ x y, 0 ≤ P x y) (hP2 : ∀ x, ∑ y, P x y = 1)
    (hirr : ∀ x y, ∃ n : ℕ, 0 < (P ^ n) x y)
    (K : Ω → ℝ) (hharm : ∀ x, K x = ∑ z, P x z * K z) (a b : Ω) : K a = K b := by
  obtain ⟨x0, -, hx0⟩ := Finset.exists_max_image Finset.univ K
    ⟨Classical.arbitrary Ω, Finset.mem_univ _⟩
  set m := K x0 with hm
  have hmax : ∀ z, K z ≤ m := fun z => hx0 z (Finset.mem_univ z)
  have hAclosed : ∀ x, K x = m → ∀ z, 0 < P x z → K z = m := by
    intro x hx z hz
    exact prop_step P hP1 hP2 K m hmax x hx (hharm x) z hz
  have hall : ∀ y, K y = m := by
    intro y
    obtain ⟨n, hn⟩ := hirr x0 y
    exact reach_closed P hP1 (fun w => K w = m) hAclosed x0 rfl n y hn
  rw [hall a, hall b]

/-- Hitting-time triangle inequality: `g a b ≤ g a c + g c b`. -/
theorem tri_ineq {Ω : Type*} [Fintype Ω] [Nonempty Ω] [DecidableEq Ω]
    (P : Matrix Ω Ω ℝ) (hP1 : ∀ x y, 0 ≤ P x y) (hP2 : ∀ x, ∑ y, P x y = 1)
    (hirr : ∀ x y, ∃ n : ℕ, 0 < (P ^ n) x y)
    (g : Ω → Ω → ℝ) (hg : ∀ x y, g x y = if x = y then 0 else 1 + ∑ z, P x z * g z y)
    (gnn : ∀ x y, 0 ≤ g x y) (a b c : Ω) : g a b ≤ g a c + g c b := by
  by_cases hbc : b = c
  · subst hbc
    have h0 : g b b = 0 := by rw [hg b b, if_pos rfl]
    have := gnn a b; nlinarith [h0]
  set θ : Ω → ℝ := fun x => g x b - g x c with hθ
  have hharm : ∀ x, x ≠ b → x ≠ c → θ x = ∑ z, P x z * θ z := by
    intro x hxb hxc
    have gb : g x b = 1 + ∑ z, P x z * g z b := by rw [hg x b, if_neg hxb]
    have gc : g x c = 1 + ∑ z, P x z * g z c := by rw [hg x c, if_neg hxc]
    simp only [hθ]; rw [gb, gc]
    have : (∑ z, P x z * (g z b - g z c)) = (∑ z, P x z * g z b) - ∑ z, P x z * g z c := by
      rw [← Finset.sum_sub_distrib]; apply Finset.sum_congr rfl; intro z _; ring
    rw [this]; ring
  obtain ⟨x0, -, hx0⟩ := Finset.exists_max_image Finset.univ θ
    ⟨Classical.arbitrary Ω, Finset.mem_univ _⟩
  set m := θ x0 with hm
  have hmax : ∀ z, θ z ≤ m := fun z => hx0 z (Finset.mem_univ z)
  have hθc : θ c = g c b := by simp only [hθ]; rw [hg c c, if_pos rfl]; ring
  have hθb : θ b = - g b c := by simp only [hθ]; rw [hg b b, if_pos rfl]; ring
  have hmeq : m = θ c := by
    by_cases hcA : θ c = m
    · exact hcA.symm
    · by_cases hbA : θ b = m
      · exfalso
        have h1 : m ≤ 0 := by rw [← hbA, hθb]; linarith [gnn b c]
        have h2 : 0 ≤ θ c := by rw [hθc]; exact gnn c b
        have h3 : θ c ≤ m := hmax c
        apply hcA; linarith
      · exfalso
        have hAclosed : ∀ x, θ x = m → ∀ z, 0 < P x z → θ z = m := by
          intro x hx z hz
          have hxb : x ≠ b := fun h => hbA (h ▸ hx)
          have hxc : x ≠ c := fun h => hcA (h ▸ hx)
          exact prop_step P hP1 hP2 θ m hmax x hx (hharm x hxb hxc) z hz
        obtain ⟨n, hn⟩ := hirr x0 c
        exact hcA (reach_closed P hP1 (fun w => θ w = m) hAclosed x0 rfl n c hn)
  have hθa : θ a ≤ θ c := by rw [← hmeq]; exact hmax a
  rw [hθc] at hθa
  simp only [hθ] at hθa
  linarith

/-- The core inequality: every hitting time is at most `2 * M`. -/
theorem core_bound {Ω : Type*} [Fintype Ω] [Nonempty Ω] [DecidableEq Ω]
    (P : Matrix Ω Ω ℝ) (hP1 : ∀ x y, 0 ≤ P x y) (hP2 : ∀ x, ∑ y, P x y = 1)
    (hirr : ∀ x y, ∃ n : ℕ, 0 < (P ^ n) x y)
    (g : Ω → Ω → ℝ) (hg : ∀ x y, g x y = if x = y then 0 else 1 + ∑ z, P x z * g z y)
    (π : Ω → ℝ) (hpi1 : ∀ x, 0 ≤ π x) (hpi2 : ∑ x, π x = 1)
    (hpi3 : ∀ y, ∑ x, π x * P x y = π y)
    (M : ℝ) (hM : ∀ y : Ω, (∑ x, π x * g x y) ≤ M) (a b : Ω) :
    g a b ≤ 2 * M := by
  have gnn : ∀ x y, 0 ≤ g x y := fun x y => g_nonneg P hP1 hP2 g hg y x
  have hRT : ∀ w, π w * (1 + ∑ z, P w z * g z w) = 1 :=
    fun w => returnTime_identity P g hg π hpi2 hpi3 w
  have tri : ∀ x y z, g x y ≤ g x z + g z y :=
    fun x y z => tri_ineq P hP1 hP2 hirr g hg gnn x y z
  have Kconst : ∀ x y, (∑ c, π c * g x c) = (∑ c, π c * g y c) := by
    intro x y
    exact K_const P hP1 hP2 hirr (fun x => ∑ c, π c * g x c)
      (fun x => K_harmonic P hP1 hP2 g hg π hpi1 hpi2 hRT x) x y
  -- Step 1: g a b ≤ K a + Z b
  have hgab : g a b ≤ (∑ c, π c * g a c) + (∑ c, π c * g c b) := by
    have h1 : g a b = ∑ c, π c * g a b := by rw [← Finset.sum_mul, hpi2, one_mul]
    rw [h1, ← Finset.sum_add_distrib]
    apply Finset.sum_le_sum; intro c _
    have htri := tri a b c
    have : π c * g a b ≤ π c * (g a c + g c b) := mul_le_mul_of_nonneg_left htri (hpi1 c)
    nlinarith [this]
  -- Step 2: Z b ≤ M
  have hZb : (∑ c, π c * g c b) ≤ M := hM b
  -- Step 3: K a ≤ M
  have hKa : (∑ c, π c * g a c) ≤ M := by
    have hka_eq : (∑ c, π c * g a c) = ∑ x, π x * (∑ c, π c * g x c) := by
      have : (∑ x, π x * (∑ c, π c * g x c)) = ∑ x, π x * (∑ c, π c * g a c) := by
        apply Finset.sum_congr rfl; intro x _; rw [Kconst x a]
      rw [this, ← Finset.sum_mul, hpi2, one_mul]
    rw [hka_eq]
    have swap : (∑ x, π x * (∑ c, π c * g x c)) = ∑ c, π c * (∑ x, π x * g x c) := by
      have l1 : (∑ x, π x * (∑ c, π c * g x c)) = ∑ x, ∑ c, π x * (π c * g x c) := by
        apply Finset.sum_congr rfl; intro x _; rw [Finset.mul_sum]
      have l2 : (∑ c, π c * (∑ x, π x * g x c)) = ∑ c, ∑ x, π c * (π x * g x c) := by
        apply Finset.sum_congr rfl; intro c _; rw [Finset.mul_sum]
      rw [l1, l2, Finset.sum_comm]
      apply Finset.sum_congr rfl; intro c _; apply Finset.sum_congr rfl; intro x _; ring
    rw [swap]
    calc (∑ c, π c * (∑ x, π x * g x c)) ≤ ∑ c, π c * M := by
              apply Finset.sum_le_sum; intro c _
              exact mul_le_mul_of_nonneg_left (hM c) (hpi1 c)
      _ = M := by rw [← Finset.sum_mul, hpi2, one_mul]
  linarith

end Q361Helpers

open Auto in
theorem q361_hitting_bound {Ω : Type*} [Fintype Ω] [Nonempty Ω] [DecidableEq Ω]
    (P : Matrix Ω Ω ℝ) (hP : IsStochastic P) (hirr : IsIrreducible P)
    (g : Ω → Ω → ℝ) (hg : IsHittingSolution P g)
    (π : Ω → ℝ) (hπ : IsStationary P π) :
    (⨆ p : Ω × Ω, g p.1 p.2) ≤ 2 * ⨆ w, ∑ x, π x * g x w := by
  obtain ⟨hP1, hP2⟩ := hP
  obtain ⟨hpi1, hpi2, hpi3⟩ := hπ
  have hreach : ∀ x y, ∃ n : ℕ, 0 < (P ^ n) x y := by
    intro x y
    obtain ⟨n, _, hn⟩ := hirr x y
    exact ⟨n, hn⟩
  set M := ⨆ w, ∑ x, π x * g x w with hMdef
  have hbddM : BddAbove (Set.range (fun w => ∑ x, π x * g x w)) := Finite.bddAbove_range _
  have hZle : ∀ y : Ω, (∑ x, π x * g x y) ≤ M := fun y => le_ciSup hbddM y
  apply ciSup_le
  intro p
  exact Q361Helpers.core_bound P hP1 hP2 hreach g hg π hpi1 hpi2 hpi3 M hZle p.1 p.2
\end{lstlisting}

\begin{figure}[htbp]
    \centering
    \includegraphics[width=0.9\textwidth]{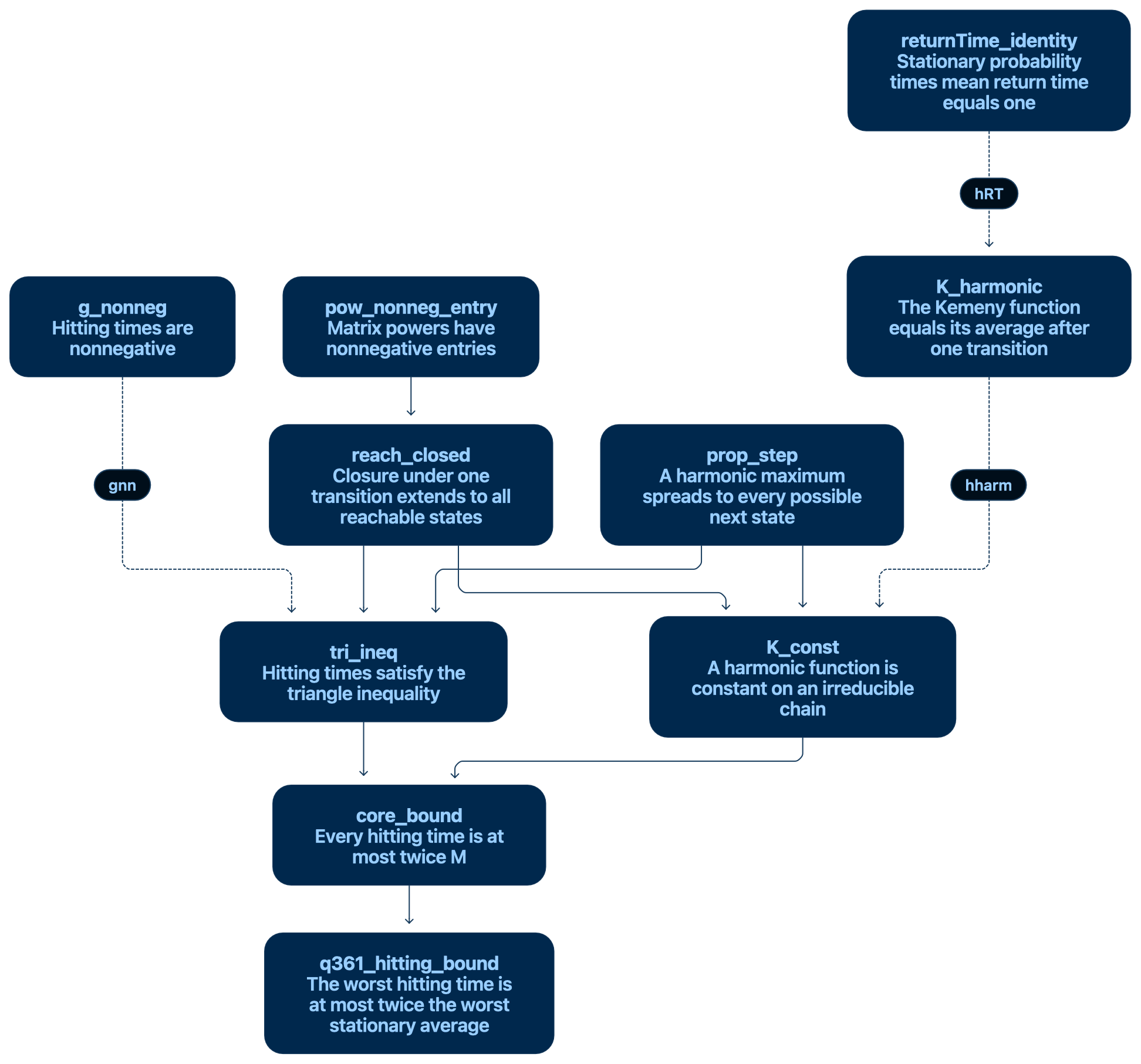}
    \caption{Main logical dependencies in the generated proof of Q361.
Shared maximum-principle lemmas establish Kemeny constancy and the
hitting-time triangle inequality, which are combined by stationary
averaging to prove the bound. Nodes give Lean theorem names and their
mathematical roles; dashed arrows indicate results supplied as
hypotheses to later lemmas.}
\label{fig:q361-proof-dependencies}
\end{figure}

\subsection{A direct proof example - Opus Prover}
\label{sec:q222-direct-proof}


We illustrate Q222 through its natural-language statement, Lean abstraction, and an agent-generated proof obtained in a separate run lasting less than 15 minutes.

\paragraph{Natural-language statement.}
Suppose that P is irreducible. Show that there exists a probability measure $\pi$ such that $\pi = \pi P$ and $\pi(x) > 0$ for all $x \in \Omega$.

\paragraph{Formalization.}
The Lean statement represents the chain by a real transition matrix $P$
on a finite, nonempty state space $\Omega$. The assumptions
\texttt{IsStochastic} and \texttt{IsIrreducible} specify nonnegative
entries, unit row sums, and reachability through positive entries of
matrix powers. A stationary probability measure is represented by a
function $\pi:\Omega\to\mathbb{R}$ satisfying
\[
  \pi(x)\geq 0,
  \qquad
  \sum_{x\in\Omega}\pi(x)=1,
  \qquad
  \sum_{x\in\Omega}\pi(x)P(x,y)=\pi(y).
\]

\begin{lstlisting}
import Mathlib
import Leanproj.Basic
open MeasureTheory ProbabilityTheory Filter
open scoped ENNReal NNReal Topology Matrix
open Auto
variable {Ω : Type*} {m0 : MeasurableSpace Ω}

/-- The pushforward of a distribution `v` by the transition matrix `P`: `(vP) y = ∑ x, v x P x y`. -/
private def stepMap {Ω : Type*} [Fintype Ω] (P : Matrix Ω Ω ℝ) (v : Ω → ℝ) : Ω → ℝ :=
  fun y => ∑ x, v x * P x y

/-- Cesàro averages of the iterates of `stepMap P` started at `v0`. -/
private noncomputable def cesaro {Ω : Type*} [Fintype Ω] (P : Matrix Ω Ω ℝ) (v0 : Ω → ℝ)
    (N : ℕ) : Ω → ℝ :=
  fun y => (1 / (N + 1 : ℝ)) * ∑ n ∈ Finset.range (N + 1), ((stepMap P)^[n] v0) y

/-- `stepMap P` maps the standard simplex into itself. -/
private lemma stepMap_mem {Ω : Type*} [Fintype Ω] {P : Matrix Ω Ω ℝ}
    (hP0 : ∀ x y, 0 ≤ P x y) (hP1 : ∀ x, ∑ y, P x y = 1)
    {v : Ω → ℝ} (hv : v ∈ stdSimplex ℝ Ω) : stepMap P v ∈ stdSimplex ℝ Ω := by
  refine ⟨fun y => ?_, ?_⟩
  · show 0 ≤ ∑ x, v x * P x y
    exact Finset.sum_nonneg (fun x _ => mul_nonneg (hv.1 x) (hP0 x y))
  · show ∑ y, ∑ x, v x * P x y = 1
    rw [Finset.sum_comm]
    calc ∑ x, ∑ y, v x * P x y = ∑ x, v x * ∑ y, P x y := by
            apply Finset.sum_congr rfl; intro x _; rw [Finset.mul_sum]
      _ = ∑ x, v x * 1 := by apply Finset.sum_congr rfl; intro x _; rw [hP1 x]
      _ = ∑ x, v x := by apply Finset.sum_congr rfl; intro x _; rw [mul_one]
      _ = 1 := hv.2

/-- Entries of powers of a nonnegative matrix are nonnegative. -/
private lemma pow_nonneg_entries {Ω : Type*} [Fintype Ω] [DecidableEq Ω] {P : Matrix Ω Ω ℝ}
    (hP0 : ∀ x y, 0 ≤ P x y) : ∀ n x y, 0 ≤ (P ^ n) x y := by
  intro n
  induction n with
  | zero => intro x y; rw [pow_zero]; by_cases h : x = y <;> simp [Matrix.one_apply, h]
  | succ n ih =>
    intro x y
    rw [pow_succ, Matrix.mul_apply]
    exact Finset.sum_nonneg (fun z _ => mul_nonneg (ih x z) (hP0 z y))

/-- A stationary distribution is stationary for every power of `P`. -/
private lemma stationary_pow {Ω : Type*} [Fintype Ω] [DecidableEq Ω] {P : Matrix Ω Ω ℝ}
    {π : Ω → ℝ} (hstat : ∀ y, ∑ x, π x * P x y = π y) :
    ∀ n y, ∑ x, π x * (P ^ n) x y = π y := by
  intro n
  induction n with
  | zero =>
    intro y
    rw [pow_zero]
    simp [Matrix.one_apply, Finset.sum_ite_eq']
  | succ n ih =>
    intro y
    calc ∑ x, π x * (P ^ (n + 1)) x y
        = ∑ x, ∑ z, π x * ((P ^ n) x z * P z y) := by
          apply Finset.sum_congr rfl; intro x _
          rw [pow_succ, Matrix.mul_apply, Finset.mul_sum]
      _ = ∑ z, ∑ x, π x * ((P ^ n) x z * P z y) := Finset.sum_comm
      _ = ∑ z, (∑ x, π x * (P ^ n) x z) * P z y := by
          apply Finset.sum_congr rfl; intro z _
          rw [Finset.sum_mul]; apply Finset.sum_congr rfl; intro x _; ring
      _ = ∑ z, π z * P z y := by
          apply Finset.sum_congr rfl; intro z _; rw [ih z]
      _ = π y := hstat y

theorem q222_stationary_existence_positive {Ω : Type*} [Fintype Ω] [DecidableEq Ω] [Nonempty Ω]
    (P : Matrix Ω Ω ℝ) (hP : IsStochastic P) (hirr : IsIrreducible P) :
    ∃ π : Ω → ℝ, IsStationary P π ∧ ∀ x, 0 < π x := by
  obtain ⟨hP0, hP1⟩ := hP
  -- Choose a starting point in the simplex.
  obtain ⟨v0, hv0⟩ : (stdSimplex ℝ Ω).Nonempty := Set.nonempty_coe_sort.mp inferInstance
  -- The iterates of `stepMap P` stay in the simplex.
  have hiter : ∀ n, (stepMap P)^[n] v0 ∈ stdSimplex ℝ Ω := by
    intro n
    induction n with
    | zero => simpa only [Function.iterate_zero_apply] using hv0
    | succ n ih => rw [Function.iterate_succ_apply']; exact stepMap_mem hP0 hP1 ih
  -- The Cesàro averages stay in the simplex.
  have hbmem : ∀ N, cesaro P v0 N ∈ stdSimplex ℝ Ω := by
    intro N
    refine ⟨fun y => ?_, ?_⟩
    · show 0 ≤ (1 / (N + 1 : ℝ)) * ∑ n ∈ Finset.range (N + 1), ((stepMap P)^[n] v0) y
      exact mul_nonneg (by positivity) (Finset.sum_nonneg (fun n _ => (hiter n).1 y))
    · show ∑ y, (1 / (N + 1 : ℝ)) * ∑ n ∈ Finset.range (N + 1), ((stepMap P)^[n] v0) y = 1
      rw [← Finset.mul_sum, Finset.sum_comm]
      rw [Finset.sum_congr rfl (fun n (_ : n ∈ Finset.range (N + 1)) => (hiter n).2)]
      rw [Finset.sum_const, Finset.card_range]
      simp only [nsmul_eq_mul, mul_one, Nat.cast_add, Nat.cast_one]
      exact one_div_mul_cancel (by positivity)
  -- Key algebraic identity: `stepMap` moves the Cesàro average by a telescoping tail.
  have hdiff : ∀ N y, stepMap P (cesaro P v0 N) y - cesaro P v0 N y
      = (1 / (N + 1 : ℝ)) * (((stepMap P)^[N + 1] v0) y - v0 y) := by
    intro N y
    have hTb : stepMap P (cesaro P v0 N) y
        = (1 / (N + 1 : ℝ)) * ∑ n ∈ Finset.range (N + 1), ((stepMap P)^[n + 1] v0) y := by
      show ∑ x, cesaro P v0 N x * P x y = _
      have e1 : ∀ x, cesaro P v0 N x * P x y
          = (1 / (N + 1 : ℝ)) * ∑ n ∈ Finset.range (N + 1), ((stepMap P)^[n] v0) x * P x y := by
        intro x
        show ((1 / (N + 1 : ℝ)) * ∑ n ∈ Finset.range (N + 1), ((stepMap P)^[n] v0) x) * P x y = _
        rw [mul_assoc, Finset.sum_mul]
      rw [Finset.sum_congr rfl (fun x (_ : x ∈ Finset.univ) => e1 x), ← Finset.mul_sum]
      congr 1
      rw [Finset.sum_comm]
      apply Finset.sum_congr rfl; intro n _
      show ∑ x, ((stepMap P)^[n] v0) x * P x y = ((stepMap P)^[n + 1] v0) y
      rw [Function.iterate_succ_apply']; rfl
    have hbNy : cesaro P v0 N y
        = (1 / (N + 1 : ℝ)) * ∑ n ∈ Finset.range (N + 1), ((stepMap P)^[n] v0) y := rfl
    rw [hTb, hbNy, ← mul_sub]
    congr 1
    rw [← Finset.sum_sub_distrib, Finset.sum_range_sub (fun n => ((stepMap P)^[n] v0) y)]
    simp [Function.iterate_zero_apply]
  -- Hence `stepMap` of the average minus the average tends to `0`.
  have htend0 : ∀ y, Filter.Tendsto (fun N => stepMap P (cesaro P v0 N) y - cesaro P v0 N y)
      Filter.atTop (nhds 0) := by
    intro y
    refine squeeze_zero_norm (fun N => ?_) tendsto_one_div_add_atTop_nhds_zero_nat
    rw [hdiff N y, norm_mul]
    have ha := mem_Icc_of_mem_stdSimplex (hiter (N + 1)) y
    have hb := mem_Icc_of_mem_stdSimplex hv0 y
    rw [Set.mem_Icc] at ha hb
    have hcnn : (0 : ℝ) ≤ 1 / ((N : ℝ) + 1) := by positivity
    rw [Real.norm_of_nonneg hcnn]
    have hbound : ‖((stepMap P)^[N + 1] v0) y - v0 y‖ ≤ 1 := by
      rw [Real.norm_eq_abs, abs_le]
      refine ⟨?_, ?_⟩ <;> linarith [ha.1, ha.2, hb.1, hb.2]
    calc 1 / ((N : ℝ) + 1) * ‖((stepMap P)^[N + 1] v0) y - v0 y‖
        ≤ 1 / ((N : ℝ) + 1) * 1 := mul_le_mul_of_nonneg_left hbound hcnn
      _ = 1 / ((N : ℝ) + 1) := mul_one _
  -- Extract a convergent subsequence of the averages; its limit is stationary.
  obtain ⟨π, hπsimplex, φ, hφmono, hφtend⟩ := (isCompact_stdSimplex ℝ Ω).tendsto_subseq hbmem
  have hπcoord : ∀ x, Filter.Tendsto (fun N => cesaro P v0 (φ N) x) Filter.atTop (nhds (π x)) :=
    fun x => tendsto_pi_nhds.mp hφtend x
  have key : ∀ y, ∑ x, π x * P x y = π y := by
    intro y
    have hL1 : Filter.Tendsto (fun N => ∑ x, cesaro P v0 (φ N) x * P x y) Filter.atTop
        (nhds (∑ x, π x * P x y)) :=
      tendsto_finset_sum _ (fun x _ => (hπcoord x).mul_const (P x y))
    have hL2 : Filter.Tendsto (fun N => cesaro P v0 (φ N) y) Filter.atTop (nhds (π y)) := hπcoord y
    have hLdiff : Filter.Tendsto
        (fun N => (∑ x, cesaro P v0 (φ N) x * P x y) - cesaro P v0 (φ N) y)
        Filter.atTop (nhds ((∑ x, π x * P x y) - π y)) := hL1.sub hL2
    have hL0 : Filter.Tendsto
        (fun N => (∑ x, cesaro P v0 (φ N) x * P x y) - cesaro P v0 (φ N) y)
        Filter.atTop (nhds 0) := (htend0 y).comp hφmono.tendsto_atTop
    have huniq := tendsto_nhds_unique hLdiff hL0
    linarith [huniq]
  refine ⟨π, ⟨hπsimplex.1, hπsimplex.2, key⟩, ?_⟩
  -- Strict positivity via irreducibility.
  intro x0
  obtain ⟨a, ha⟩ : ∃ a, 0 < π a := by
    by_contra h
    push_neg at h
    have h0 : ∑ x, π x = 0 :=
      Finset.sum_eq_zero (fun a _ => le_antisymm (h a) (hπsimplex.1 a))
    rw [hπsimplex.2] at h0
    exact one_ne_zero h0
  obtain ⟨n, _, hn2⟩ := hirr a x0
  have hpow := stationary_pow key n x0
  rw [← hpow]
  refine Finset.sum_pos'
    (fun x _ => mul_nonneg (hπsimplex.1 x) (pow_nonneg_entries hP0 n x x0)) ?_
  exact ⟨a, Finset.mem_univ a, mul_pos ha hn2⟩

\end{lstlisting}

\begin{figure}[htbp]
     \centering
     \includegraphics[width=0.8\textwidth]{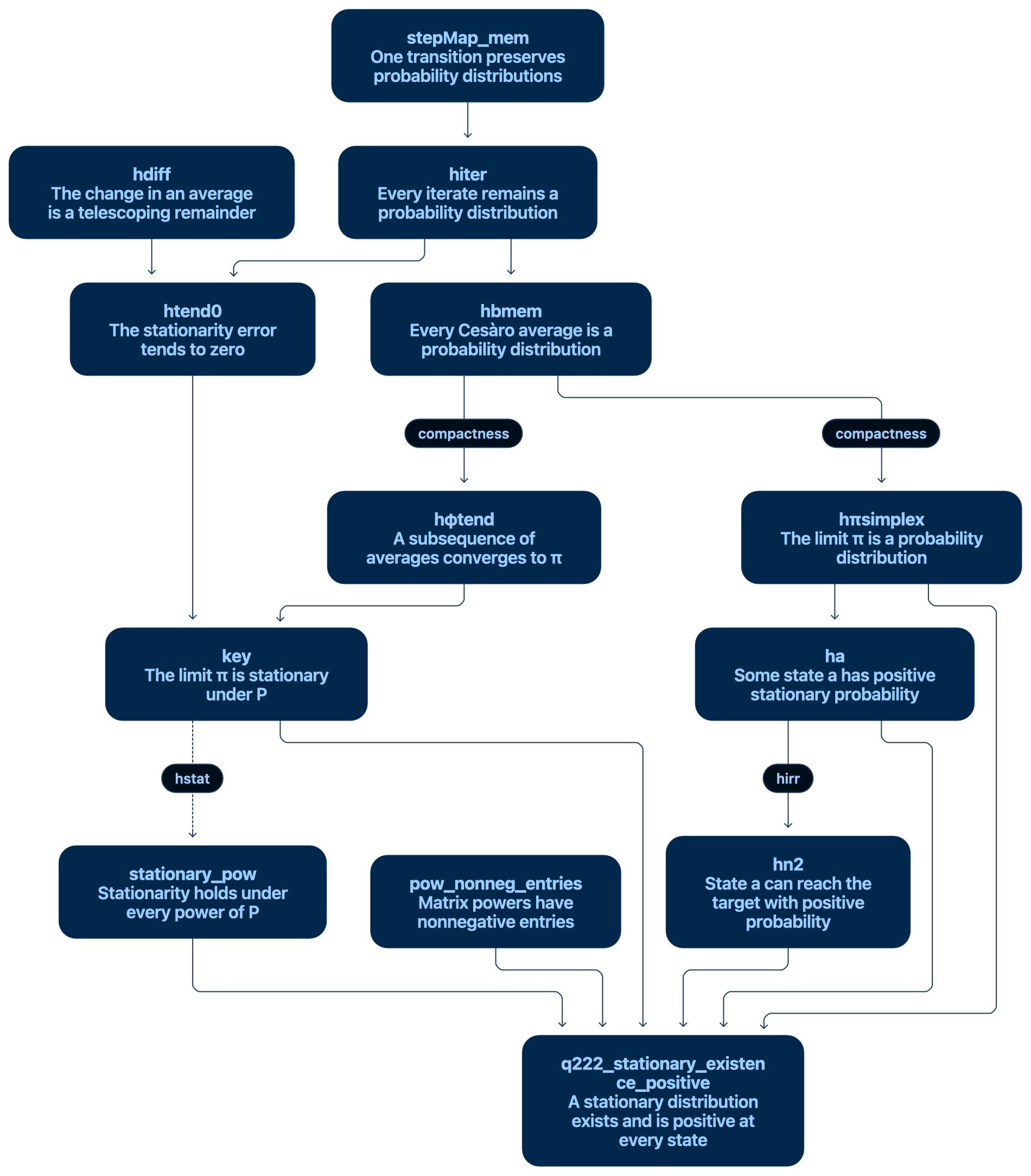}
\caption{Main logical dependencies in the generated proof of Q222.
A telescoping identity and compactness establish a stationary
subsequential limit of the Ces\`aro averages. Irreducibility and
stationarity under matrix powers then yield strict positivity.
Nodes include helper lemmas and local proof facts; edge labels
identify supplied hypotheses or supporting arguments.}
\label{fig:q222-proof-dependencies}
\end{figure}

\subsection{An abstracted proof example - Opus Prover}
\label{sec:q128-abstracted-proof}


We illustrate Q128 through its natural-language statement, Lean abstraction, and an agent-generated proof obtained in a separate run lasting less than 15 minutes.

\paragraph{Natural-language statement.}
Let $B_\mu(t) =  \mu t + \sigma B(t)$ be a Brownian motion with drift. Show that $B_\mu (t) - \mu t$ is a martingale, and that $(B_\mu(t) - \mu t)^2 - \sigma^2 t$ is a martingale.

\paragraph{Formalization.}
The Lean statement uses a real-valued process $B$ on a probability space
with measure $\mu$ and filtration $\mathcal{F}$. Adaptedness and
integrability of $B_t$ and $B_t^2$ are explicit assumptions. For
$0\leq s\leq t$, \texttt{HasLaw} specifies
$B_t-B_s\sim\mathcal{N}(0,t-s)$, and \texttt{Indep} specifies independence
of this increment from $\mathcal{F}_s$. Lean uses $c$ for the drift and
$\mu$ for the probability measure. Writing $X_t=ct+\sigma B_t$, the
conclusion consists of the almost-sure conditional-expectation identities
\[
  \begin{aligned}
    \mathbb{E}_{\mu}[X_t-ct\mid\mathcal{F}_s]
      &= X_s-cs,\\
    \mathbb{E}_{\mu}[(X_t-ct)^2-\sigma^2t\mid\mathcal{F}_s]
      &= (X_s-cs)^2-\sigma^2s.
  \end{aligned}
\]
The increment properties are supplied as hypotheses; the proof derives
the two identities from these properties.

\begin{lstlisting}
import Mathlib
import Leanproj.Basic
open MeasureTheory ProbabilityTheory Filter
open scoped ENNReal NNReal Topology Matrix
open Auto
variable {Ω : Type*} {m0 : MeasurableSpace Ω}

theorem q128_drifted_brownian_martingale (μ : Measure Ω) [IsProbabilityMeasure μ]
    ( ℱ : Filtration ℝ m0) (B : ℝ → Ω → ℝ) (c σ : ℝ)
    (hadap : Adapted ℱ B)
    (hint : ∀ t, Integrable (B t) μ)
    (hint2 : ∀ t, Integrable (fun ω => (B t ω) ^ 2) μ)
    (hB0 : ∀ᵐ ω ∂μ, B 0 ω = 0)
    (hincr : ∀ s t : ℝ, 0 ≤ s → s ≤ t →
      HasLaw (fun ω => B t ω - B s ω) (gaussianReal 0 (t - s).toNNReal) μ)
    (hpast : ∀ s t : ℝ, 0 ≤ s → s ≤ t →
      Indep (MeasurableSpace.comap (fun ω => B t ω - B s ω) inferInstance) ( ℱ s) μ) :
    (∀ s t : ℝ, 0 ≤ s → s ≤ t →
        μ[(fun ω => (c * t + σ * B t ω) - c * t) | ( ℱ s)]
          =ᵐ[μ] (fun ω => (c * s + σ * B s ω) - c * s)) ∧
    (∀ s t : ℝ, 0 ≤ s → s ≤ t →
        μ[(fun ω => ((c * t + σ * B t ω) - c * t) ^ 2 - σ ^ 2 * t) | ( ℱ s)]
          =ᵐ[μ] (fun ω => ((c * s + σ * B s ω) - c * s) ^ 2 - σ ^ 2 * s)) := by
  have hsq_int : ∀ v : NNReal, ∫ x, x ^ 2 ∂(gaussianReal (0:ℝ) v) = (v : ℝ) := by
    intro v
    have hv := variance_of_integral_eq_zero (μ := gaussianReal (0:ℝ) v) (X := fun x => x)
      measurable_id.aemeasurable (by simpa using integral_id_gaussianReal (μ := (0:ℝ)) (v := v))
    rw [variance_fun_id_gaussianReal] at hv
    simpa using hv.symm
  have main : ∀ s t : ℝ, 0 ≤ s → s ≤ t →
      (μ[B t | ℱ s] =ᵐ[μ] B s) ∧
      (μ[fun ω => (B t ω) ^ 2 | ℱ s] =ᵐ[μ] fun ω => (B s ω) ^ 2 + (t - s)) := by
    intro s t hs hst
    have hst0 : (0:ℝ) ≤ t - s := by linarith
    have hBt : Measurable (B t) := (hadap t).mono ( ℱ.le t) le_rfl
    have hBs : Measurable (B s) := (hadap s).mono ( ℱ.le s) le_rfl
    have hg_meas : Measurable (fun ω => B t ω - B s ω) := hBt.sub hBs
    have hg_int : Integrable (fun ω => B t ω - B s ω) μ := (hint t).sub (hint s)
    have hindep := hpast s t hs hst
    have hle1 : MeasurableSpace.comap (fun ω => B t ω - B s ω) inferInstance ≤ m0 :=
      hg_meas.comap_le
    have hle2 : ℱ s ≤ m0 := ℱ.le s
    have hsm : StronglyMeasurable[MeasurableSpace.comap (fun ω => B t ω - B s ω) inferInstance]
        (fun ω => B t ω - B s ω) := (measurable_iff_comap_le.2 le_rfl).stronglyMeasurable
    have hEg : μ[fun ω => B t ω - B s ω] = 0 := by
      rw [(hincr s t hs hst).integral_eq, integral_id_gaussianReal]
    have hcz : μ[fun ω => B t ω - B s ω | ℱ s] =ᵐ[μ] fun _ => (0 : ℝ) := by
      have h := condExp_indep_eq hle1 hle2 hsm hindep
      rw [hEg] at h; exact h
    have hmart : μ[B t | ℱ s] =ᵐ[μ] B s := by
      have hcond_Bs : μ[B s | ℱ s] = B s :=
        condExp_of_stronglyMeasurable hle2 (hadap s).stronglyMeasurable (hint s)
      have key : (B t) = (B s) + (fun ω => B t ω - B s ω) := by
        funext ω; show B t ω = B s ω + (B t ω - B s ω); ring
      have hadd : μ[B t | ℱ s] =ᵐ[μ]
          (μ[B s | ℱ s] + μ[fun ω => B t ω - B s ω | ℱ s]) := by
        calc μ[B t | ℱ s] = μ[(B s) + (fun ω => B t ω - B s ω) | ℱ s] :=
              congrArg (fun h => μ[h | ℱ s]) key
          _ =ᵐ[μ] _ := condExp_add (hint s) hg_int _
      filter_upwards [hadd, hcz] with ω h1 h3
      rw [h1]; simp [Pi.add_apply, hcond_Bs, h3]
    refine ⟨hmart, ?_⟩
    have hmemBt : MemLp (B t) 2 μ :=
      (memLp_two_iff_integrable_sq hBt.aestronglyMeasurable).2 (hint2 t)
    have hmemBs : MemLp (B s) 2 μ :=
      (memLp_two_iff_integrable_sq hBs.aestronglyMeasurable).2 (hint2 s)
    have hmemInc : MemLp (fun ω => B t ω - B s ω) 2 μ := hmemBt.sub hmemBs
    have hd_int : Integrable (fun ω => (B t ω - B s ω) ^ 2) μ := hmemInc.integrable_sq
    have hmem2Bs : MemLp (fun ω => 2 * B s ω) 2 μ := hmemBs.const_mul 2
    have hb_int : Integrable ((fun ω => 2 * B s ω) * (fun ω => B t ω - B s ω)) μ :=
      hmem2Bs.integrable_mul hmemInc
    have ha_int : Integrable (fun ω => (B s ω) ^ 2) μ := hint2 s
    have ea : μ[fun ω => (B s ω) ^ 2 | ℱ s] = fun ω => (B s ω) ^ 2 :=
      condExp_of_stronglyMeasurable hle2 ((hadap s).pow_const 2).stronglyMeasurable ha_int
    have eb : μ[(fun ω => 2 * B s ω) * (fun ω => B t ω - B s ω) | ℱ s] =ᵐ[μ] fun _ => (0:ℝ) := by
      have hpull := condExp_mul_of_stronglyMeasurable_left
        (((hadap s).const_mul 2).stronglyMeasurable) hb_int hg_int
      filter_upwards [hpull, hcz] with ω h1 h3
      rw [h1]; simp [Pi.mul_apply, h3]
    have hsm_d : StronglyMeasurable[MeasurableSpace.comap (fun ω => B t ω - B s ω) inferInstance]
        (fun ω => (B t ω - B s ω) ^ 2) :=
      ((measurable_iff_comap_le.2 le_rfl).pow_const 2).stronglyMeasurable
    have hEd : μ[fun ω => (B t ω - B s ω) ^ 2] = t - s := by
      have hc := (hincr s t hs hst).integral_comp (f := fun x : ℝ => x ^ 2) (by fun_prop)
      simp only [Function.comp_def] at hc
      rw [hc, hsq_int, Real.coe_toNNReal _ hst0]
    have ed : μ[fun ω => (B t ω - B s ω) ^ 2 | ℱ s] =ᵐ[μ] fun _ => (t - s) := by
      have h := condExp_indep_eq hle1 hle2 hsm_d hindep
      rw [hEd] at h; exact h
    have key2 : (fun ω => (B t ω) ^ 2) =
        (fun ω => (B s ω) ^ 2) + (fun ω => 2 * B s ω) * (fun ω => B t ω - B s ω)
          + (fun ω => (B t ω - B s ω) ^ 2) := by
      funext ω; simp only [Pi.add_apply, Pi.mul_apply]; ring
    have hcongr : μ[fun ω => (B t ω) ^ 2 | ℱ s] =
        μ[(fun ω => (B s ω) ^ 2) + (fun ω => 2 * B s ω) * (fun ω => B t ω - B s ω)
          + (fun ω => (B t ω - B s ω) ^ 2) | ℱ s] :=
      congrArg (fun h => μ[h | ℱ s]) key2
    have e2 : μ[(fun ω => (B s ω) ^ 2) + (fun ω => 2 * B s ω) * (fun ω => B t ω - B s ω) | ℱ s]
        =ᵐ[μ] (μ[fun ω => (B s ω) ^ 2 | ℱ s]
          + μ[(fun ω => 2 * B s ω) * (fun ω => B t ω - B s ω) | ℱ s]) :=
      condExp_add ha_int hb_int _
    have e1 : μ[(fun ω => (B s ω) ^ 2) + (fun ω => 2 * B s ω) * (fun ω => B t ω - B s ω)
          + (fun ω => (B t ω - B s ω) ^ 2) | ℱ s]
        =ᵐ[μ] (μ[(fun ω => (B s ω) ^ 2) + (fun ω => 2 * B s ω) * (fun ω => B t ω - B s ω) | ℱ s]
          + μ[fun ω => (B t ω - B s ω) ^ 2 | ℱ s]) :=
      condExp_add (ha_int.add hb_int) hd_int _
    rw [hcongr]
    filter_upwards [e1, e2, eb, ed] with ω h1 h2 hb hd
    rw [h1]
    simp only [Pi.add_apply] at h2 ⊢
    rw [h2, ea]
    simp only [Pi.add_apply] at hb hd ⊢
    rw [hb, hd]
    ring
  constructor
  · intro s t hs hst
    have h368 := (main s t hs hst).1
    have harg : (fun ω => (c * t + σ * B t ω) - c * t) = σ • B t := by
      funext ω; simp only [Pi.smul_apply, smul_eq_mul]; ring
    rw [harg]
    have hsmul := condExp_smul (μ := μ) σ (B t) ( ℱ s : MeasurableSpace Ω)
    filter_upwards [hsmul, h368] with ω h1 h2
    simp only [Pi.smul_apply, smul_eq_mul] at h1
    rw [h1, h2]; ring
  · intro s t hs hst
    have hquad := (main s t hs hst).2
    have harg : (fun ω => ((c * t + σ * B t ω) - c * t) ^ 2 - σ ^ 2 * t) =
        (σ ^ 2 • (fun ω => (B t ω) ^ 2)) - (fun _ => σ ^ 2 * t) := by
      funext ω; simp only [Pi.sub_apply, Pi.smul_apply, smul_eq_mul]; ring
    rw [harg]
    have hconst : μ[(fun _ => σ ^ 2 * t) | ℱ s] = fun _ => σ ^ 2 * t :=
      condExp_const ( ℱ.le s) _
    have hsmul := condExp_smul (μ := μ) (σ ^ 2) (fun ω => (B t ω) ^ 2) ( ℱ s : MeasurableSpace Ω)
    have hBt2 : MemLp (B t) 2 μ :=
      (memLp_two_iff_integrable_sq
        (((hadap t).mono ( ℱ.le t) le_rfl).aestronglyMeasurable)).2 (hint2 t)
    have hsub : μ[(σ ^ 2 • (fun ω => (B t ω) ^ 2)) - (fun _ => σ ^ 2 * t) | ℱ s]
        =ᵐ[μ] (μ[σ ^ 2 • (fun ω => (B t ω) ^ 2) | ℱ s] - μ[(fun _ => σ ^ 2 * t) | ℱ s]) :=
      condExp_sub (Integrable.smul (σ ^ 2) hBt2.integrable_sq) (integrable_const _) _
    filter_upwards [hsub, hsmul, hquad] with ω hs1 hs2 hq
    rw [hs1]
    simp only [Pi.sub_apply, hconst]
    rw [hs2]
    simp only [Pi.smul_apply, smul_eq_mul]
    rw [hq]
    ring
\end{lstlisting}

\begin{figure}[htbp]
     \centering
     \includegraphics[width=0.9\textwidth]{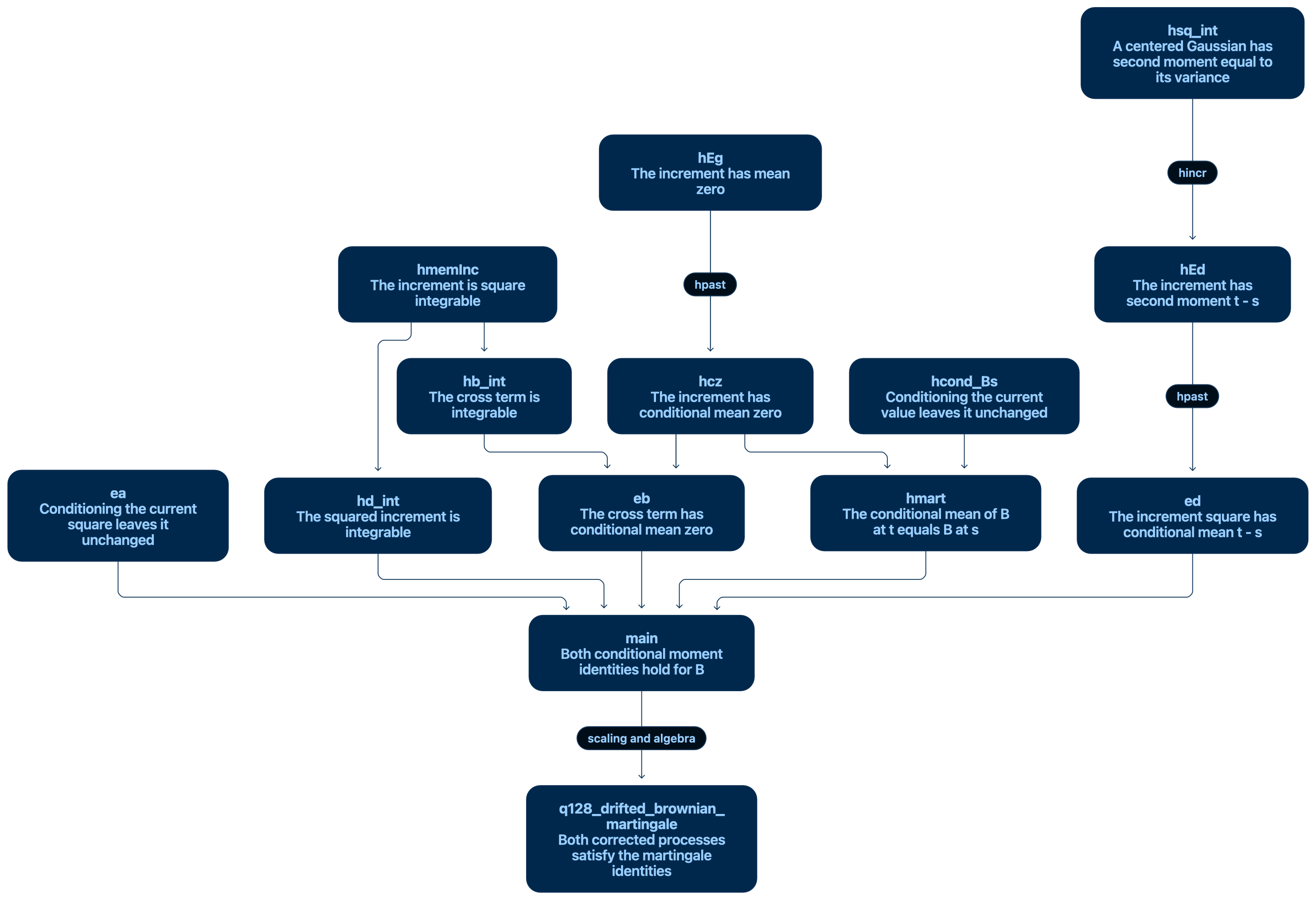}
 \caption{Main logical dependencies in the generated proof of Q128.
 Both conclusions use the zero conditional increment mean. The
 squared-process argument additionally requires the conditional second
 moment and integrability of the cross term. Deterministic scaling and
 subtraction complete the proof. Nodes name local proof facts and the
 final theorem; edge labels identify additional hypotheses or
 algebraic steps.}
 \label{fig:q128-proof-dependencies}
\end{figure}
\end{document}